\documentclass[10pt,twocolumn,letterpaper]{article}

\usepackage{iccv}
\usepackage{times}
\usepackage{epsfig}
\usepackage{graphicx}
\usepackage{amsmath}
\usepackage{amssymb}

\usepackage{soul}
\usepackage{amsthm}
\usepackage{booktabs}
\usepackage{algorithm}
\usepackage{algorithmic}
\usepackage{makecell}
\usepackage{appendix}
\usepackage{subfigure}
\usepackage[nobiblatex]{xurl}

\usepackage[pagebackref=true,breaklinks=true,letterpaper=true,colorlinks,bookmarks=false]{hyperref}

\iccvfinalcopy 

\def\iccvPaperID{8273} 

\begin{document}

\title{Incorporating Learnable Membrane Time Constant to Enhance Learning of Spiking Neural Networks}

\author{
	Wei Fang$^{1,2}$, Zhaofei Yu$^{1,2,3 *}$, Yanqi Chen$^{1,2}$, Timothée Masquelier$^{4}$, Tiejun Huang$^{1,2,3}$, Yonghong Tian$^{1,2}$\thanks{Corresponding author}\\
	$^1$Department of Computer Science and Technology, Peking University, China\\
	$^2$Peng Cheng Laboratory, China\\
	$^3$Institute for Artificial Intelligence, Peking University, China\\
	$^4$Centre de Recherche Cerveau et Cognition (CERCO), UMR5549 CNRS - Univ. Toulouse 3, France\\
	{\tt\small 
		\{fwei, yuzf12, chyq\}@pku.edu.cn,
		timothee.masquelier@cnrs.fr,
		\{tjhuang, yhtian\}@pku.edu.cn
	}
}

\maketitle
\ificcvfinal\thispagestyle{empty}\fi

\begin{abstract}
	Spiking Neural Networks (SNNs) have attracted enormous research interest due to temporal information processing capability, low power consumption, and high biological plausibility. However, the formulation of efficient and high-performance learning algorithms for SNNs is still challenging. Most existing learning methods learn weights only, and require manual tuning of the membrane-related parameters that determine the dynamics of a single spiking neuron. These parameters are typically chosen to be the same for all neurons, which limits the diversity of neurons and thus the expressiveness of the resulting SNNs. In this paper, we take inspiration from the observation that membrane-related parameters are different across brain regions, and propose a training algorithm that is capable of learning not only the synaptic weights but also the membrane time constants of SNNs. We show that incorporating learnable membrane time constants can make the network less sensitive to initial values and can speed up learning. In addition, we reevaluate the pooling methods in SNNs and find that max-pooling will not lead to significant information loss and have the advantage of low computation cost and binary compatibility. We evaluate the proposed method for image classification tasks on both traditional static
	MNIST, Fashion-MNIST, CIFAR-10 datasets, and neuromorphic N-MNIST, CIFAR10-DVS, DVS128 Gesture datasets.
	The experiment results show that the proposed method outperforms the state-of-the-art accuracy on nearly all datasets, using fewer time-steps.  
	Our codes are available at \url{https://github.com/fangwei123456/Parametric-Leaky-Integrate-and-Fire-Spiking-Neuron}.

\end{abstract}

\section{Introduction}
Spiking Neural Networks (SNNs) are viewed as the third generation of neural network models, which are closer to biological neurons in the brain \cite{maas1997networks}.
Together with neuronal and synaptic states, the importance of spike timing is also considered in SNNs.
Due to their distinctive properties, such as temporal information processing capability, low power consumption \cite{roy2019towards}, and high biological plausibility \cite{gerstner2014neuronal},
SNNs increasingly arouse researchers' great interest in recent years. 
Nevertheless, it remains challenging to formulate efficient and high-performance learning algorithms for SNNs.

\begin{figure}
	\centering
	\subfigure[Spiking neuron]{\includegraphics[width=0.125\textwidth,trim=0 0 0 0,clip]{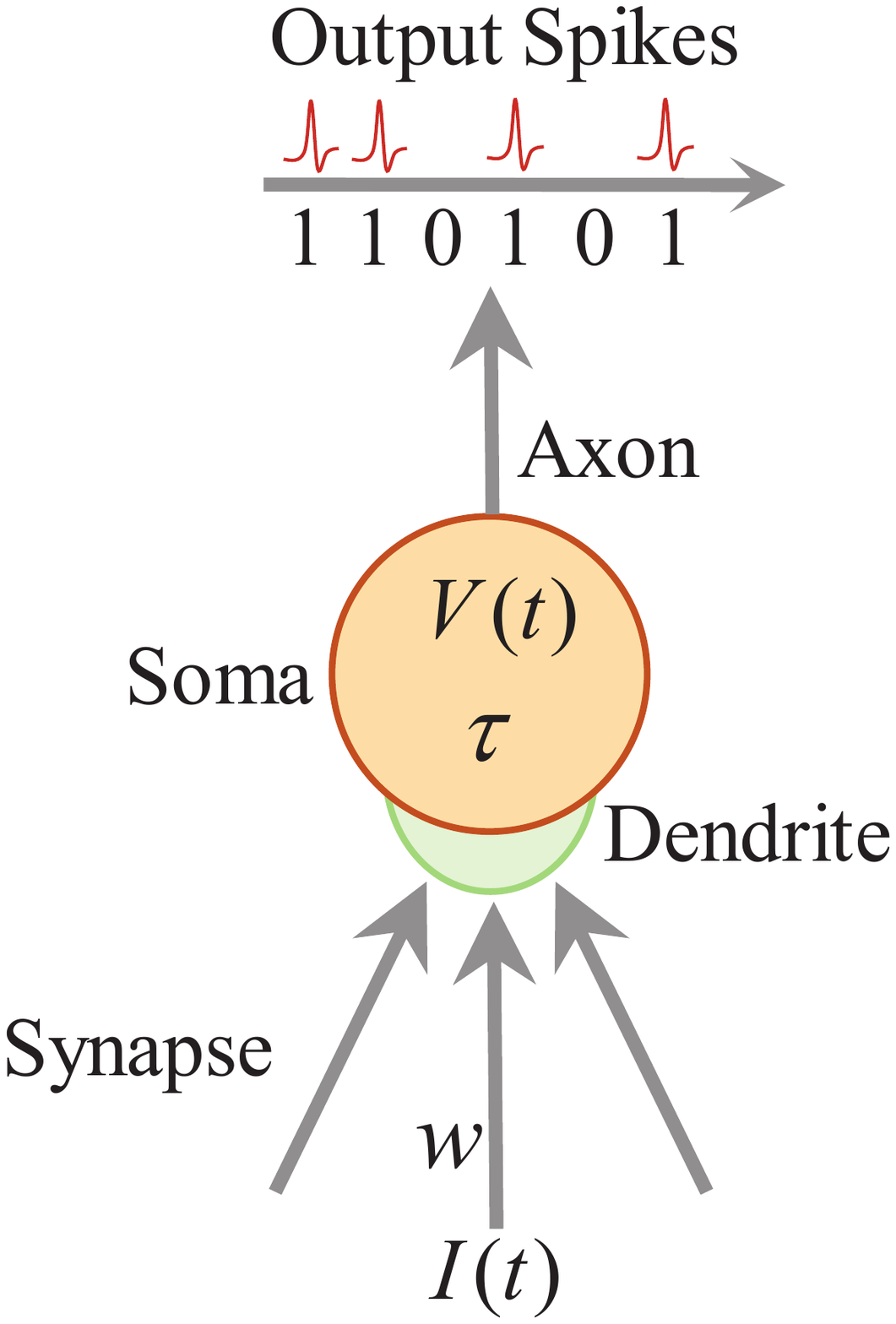}\label{figure: 1a}}
	\subfigure[The membrane potential of a LIF neuron]{\includegraphics[width=0.335\textwidth,trim=0 0 0 0,clip]{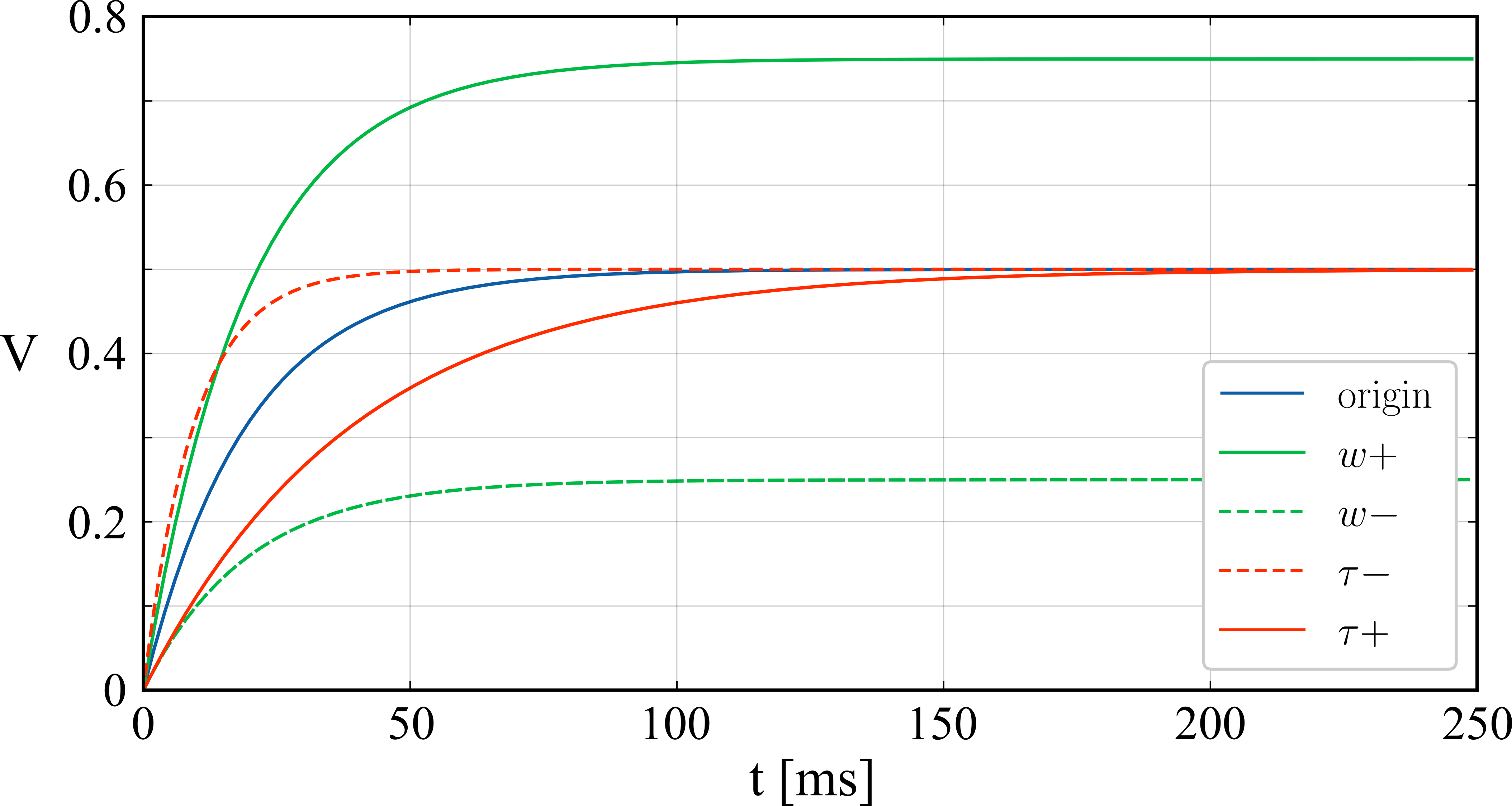}\label{figure: 1b}}
	\caption{(a) A Leaky Integrate-and-Fire (LIF)  neuron with membrane potential $V$, membrane time constant $\tau$, input $I(t)$ and synaptic weight $w$. (b) The membrane potential $V$ of the LIF neuron when constant input is received. Increasing or decreasing $\tau$ will stretch the $v=f(t)$ curve in the $t$ direction while increasing or decreasing $w$ will stretch the $v=f(t)$ curve in the $V$ direction.}
	\label{figure: 1}
\end{figure}

Generally, the learning algorithms for SNNs can be divided into unsupervised learning, supervised learning, reward-based learning, and Artificial Neural Network (ANN) to SNN conversion methodologies. Either way, we find that most existing learning methods only consider learning the synaptic-related parameters like synaptic weights and treat the membrane-related parameters as hyperparameters. These membrane-related parameters like membrane time constants, 
which determine the dynamics of a single spiking neuron, are typically chosen to be the same for all neurons. Note, however, there exist different membrane time constants for spiking neurons across brain regions \cite{mattia2002population,2019Brain,koch1996a}, which are proved to be essential for the representation of working memory and formulation of learning \cite{2006Mechanisms, Shankar2012A}. Thus simply ignoring different time constants in SNNs will limit the heterogeneity of neurons and thus the expressiveness of the resulting SNNs.

In this paper, we propose a training algorithm that is capable of learning not only the synaptic weights but also membrane time constants of SNNs. As illustrated in Fig.\,\ref{figure: 1}, we find that adjustments of the synaptic weight and the membrane time constants have different effects on neuronal dynamics. We show that incorporating learnable membrane time constants is able to enhance the learning of SNNs.

The main contributions of this paper can be summarized as follows: 
\begin{enumerate}
	\setlength{\itemsep}{0pt}
	\setlength{\parsep}{0pt}
	\setlength{\parskip}{0pt}
	\item[1)]
	We propose the backpropagation-based learning algorithm using spiking neurons with learnable membrane parameters, referred to as Parametric Leaky Integrate-and-Fire (PLIF) spiking neurons, which better represent the heterogeneity of neurons and thereby enhancing the expressiveness of the SNNs. We show that the SNNs made of PLIF neurons are more robust to initial values and can learn faster than SNNs made of neurons with a fixed time constant.
	\item[2)] We reevaluate the pooling methods in SNNs and discredit the previous conclusion that max-pooling results in significant information loss. We find that compared to average-pooling, max-pooling is able to better preserve the asynchronous characteristic of neuron firing, as well as reduce the computation cost.
	Our experiments show that the performance of max-pooling is comparable to average-pooling.
	\item[3)] We evaluate our methods on both traditional static MNIST \cite{MNIST}, Fashion-MNIST \cite{FMNIST}, CIFAR-10 \cite{CIFAR10} datasets widely used in ANNs as benchmarks, and neuromorphic N-MNIST \cite{NMNIST}, CIFAR10-DVS \cite{DVS-CIFAR10}, DVS128 Gesture \cite{Amir_2017_CVPR} datasets that focus on verifying the network's temporal information processing capability. The proposed method exceeds state-of-the-art accuracy on nearly all tested datasets, using fewer time-steps.	
\end{enumerate}

\section{Related Works}

\noindent
\textbf{Unsupervised learning of SNNs}~~The unsupervised learning methods of SNNs are based on biological plausible local learning rules, like Hebbian learning \cite{hebb1949the} and Spike-Timing-Dependent Plasticity (STDP) \cite{bi1998synaptic}. Existing approaches exploited the self-organization principle \cite{srinivasa2012self, diehl2015unsupervised, Kheradpisheh2018}, and STDP-based expectation-maximization algorithm \cite{nessler2013bayesian,guo2017hierarchical}. However, these methods are only suitable for shallow SNNs, and the performance is far below state-of-the-art ANN results.


\noindent
\textbf{Reward-based learning of SNNs}~~Reward-based learning of SNNs mimics the way the human brain learns by taking advantage of the reward or punishment signals induced by dopaminergic, serotonergic, cholinergic, or adrenergic neurons \cite{fremaux2016neuromodulated,botvinick2020deep,MOZAFARI201987}. Despite the methods that arise in reinforcement learning, like policy gradient \cite{seung2003learning,kappel2018dynamic}, temporal-difference learning \cite{potjans2011imperfect,fremaux2013reinforcement} and Q-learning \cite{botvinick2020deep}, some heuristic phenomenological models based on STDP \cite{friedrich2011spatio,yuan2019reinforcement} were proposed recently.

\noindent
\textbf{ANN to SNN conversion}~~ANN to SNN conversion (ANN2SNN) converts a trained non-spiking ANN to an SNN by using the firing rate of each spiking neuron to approximate the corresponding ReLU activation of an analog neuron \cite{hunsberger2015spiking,cao2015spiking, Bodo2017Conversion}.
It can get near lossless inference results as an ANN \cite{sengupta2019going, deng2021optimal}, but there is a trade-off between accuracy and latency. To improve accuracy, longer inference latency is needed \cite{Han_2020_CVPR}. ANN2SNN is restricted to rate-coding, which loses the processing capability in temporal tasks. As far as we know, ANN2SNN only works for static datasets, not neuromorphic datasets.


\noindent
\textbf{Supervised learning of SNNs}~~
SpikeProp \cite{BOHTE200217} was the first supervised learning method for SNNs based on backpropagation, which used a linear approximation to overcome the non-differentiable threshold-triggered firing mechanism of SNNs. Subsequent works included Tempotron \cite{tempotron}, ReSuMe \cite{ponulak2010supervised}, and SPAN \cite{mohemmed2012span}, but they could only be applied to single-layer SNNs. Recently, the surrogate gradient method was proposed and provided another solution to training multi-layer SNNs \cite{lee2016training, HM-2BP,ST-RSBP,wu2018STBP,shrestha2018slayer,lee2020enabling,10.3389/fnins.2020.00424}. It utilized surrogate derivatives to define the derivative of the threshold-triggered firing mechanism. Thus the SNNs could be optimized with gradient descent algorithms as ANNs. Zenke et al. \cite{Zenke2020.06.29.176925, Neftci2019a} systematically studied the remarkable robustness of surrogate gradient learning and showed that SNNs optimized by the surrogate gradient methods can achieve competitive performance with ANNs. Compared to ANN2SNN, the surrogate gradient method has no restrictions on simulating time-steps because it is not based on rate-coding \cite{wu2019direct, Zenke2020.06.29.176925}.

\noindent
\textbf{Spiking neurons and layers of deep SNNs}~~
Spiking neuron and layer models play an essential role in SNNs. Cheng et al. \cite{LISNN} added the lateral interactions between neighboring neurons and get better accuracy and stronger noise-robustness. Zimmer et al. \cite{zimmer2019technical} firstly adopt the learnable time constants in LIF neurons for the speech recognition task. Bellec et al. \cite{NIPS2018_7359} proposed the adaptive threshold spiking neuron to enhance computing and learning capabilities of SNNs, which was improved by \cite{yin2020effective} with learnable time constants. Rathi et al. \cite{rathi2020dietsnn} suggested using a learnable membrane leak and firing threshold to finetune SNNs converted from ANNs. Despite this, no systematic research on the effects of learning membrane time constants to SNNs has been conducted so far, which is exactly the aim of this paper. Wu et al. \cite{wu2019direct} found that normalization layers are also critical for deep SNNs and proposed Neuron Normalization (NeuNorm) to balance each neuron's firing rate to avoid severe information loss. Ledinauskas, E et al. \cite{ledinauskas2020training} firstly suggested that using Batch Normalization \cite{BN} in deep SNNs for faster convergence.


\section{Methods}
In this section, we first briefly review the Leaky Integrate-and-Fire model in Sec.~\ref{IFmodel}, and analyze the effect of synaptic weight and membrane time constant in Sec.~\ref{mtc}. The Parametric Leaky Integrate-and-Fire model and the network structure of the SNNs are then introduced in Sec.~\ref{Parametric Leaky Integrate-and-Fire Neuron} -- Sec.~\ref{networksf}.
At last, we describe the spike max-pooling and the learning algorithm of SNNs in Sec.~\ref{sectionpooling} and Sec.~\ref{sectionlearning}.

\subsection{Leaky Integrate-and-Fire model}
\label{IFmodel}

The basic computing unit of an SNN is the spiking neuron. Neuroscientists have built several spiking neuron models for describing the accurate relationships between input and output signals of the biological neuron.  The Leaky Integrate-and-Fire (LIF) model \cite{gerstner2014neuronal} is one of the simplest spiking neuron models used in SNNs. The subthreshold dynamics of the LIF neuron is defined as: 

\begin{equation}
	\tau \frac{\mathrm{d}V(t)}{\mathrm{d}t} = -(V(t) - V_{rest}) + X(t),
	\label{LIF dynamics}
\end{equation}
where $V(t)$ represents the membrane potential of the neuron at time $t$, $X(t)$ represents the input to neuron at time $t$, $\tau$ is the membrane time constant, and $V_{rest}$ is the resting potential. When the membrane potential $V(t)$ exceeds a certain threshold $V_{th}$ at time $t^f$, the neuron will elicit a spike and then the membrane potential $V(t)$ goes back to a reset value $V_{reset} < V_{th}$. The LIF neuron achieves a balance between computing cost and biological plausibility.  We set $V_{rest}=V_{reset}$ in this paper, and will not make a distinction between them in the rest of this paper.

\subsection{Function comparison of synaptic weight and membrane time constant}
\label{mtc}

In most of the previous learning algorithms for SNNs made of LIF neurons, the membrane time constant $\tau$ is regarded as a hyper-parameter and chosen to be the same for all neurons before learning. The learning of SNNs is only to optimize the synaptic weights. However, it cannot be ignored that the behavior of a spiking neuron for given inputs depends not only on the weights of connected synapses but also on the neuron's inherent dynamics controlled by the membrane time constant $\tau$.

\begin{figure}[t!]
	\centering
	{\includegraphics[width=0.45\textwidth,trim=0 0 0 0,clip]{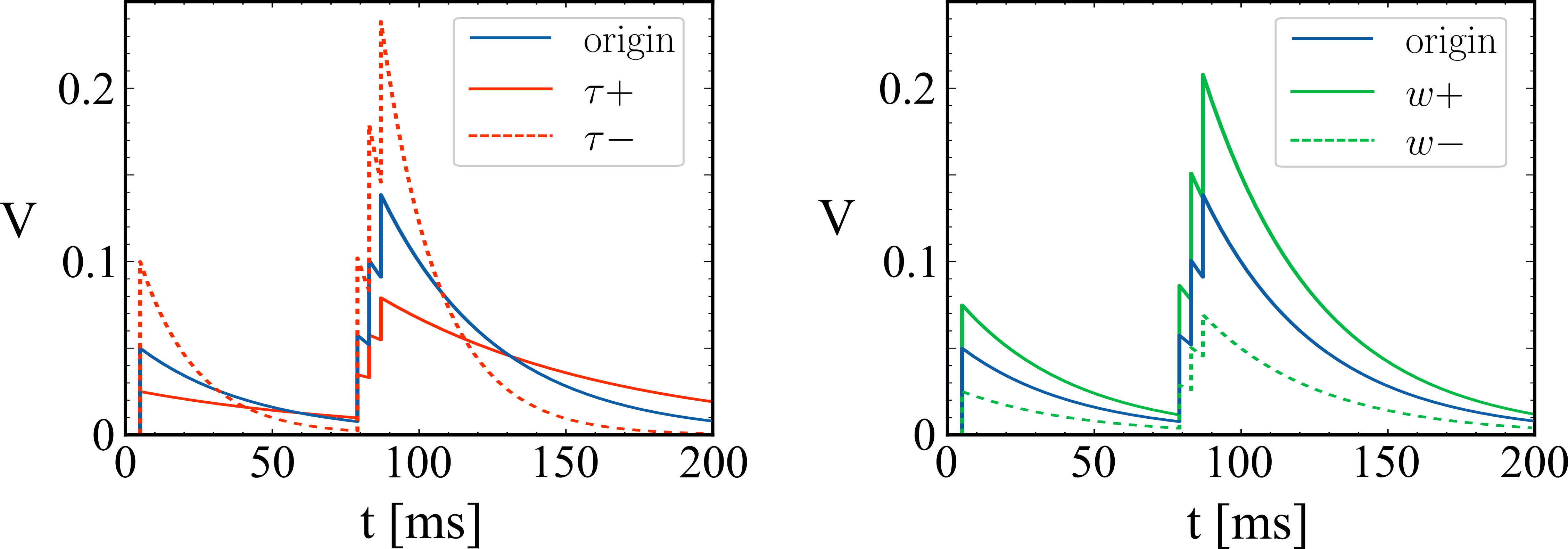}}
	\caption{The membrane potential $V$ of a LIF neuron when instant spikes at $t = 5, 80, 85, 90$ are received.}
	\label{figure: effect of tau}
\end{figure}

In order to compare the effects of synaptic weight and membrane time constant to the neuronal dynamics, we consider a simple case where the LIF neuron $z_i$ receives weighted input $X(t)=wI(t)$ from a presynaptic neuron $z_j$ (Fig.~\ref{figure: 1a}).
The rest potential $V_{rest}$ is set to $0$. When the input is constant, namely, $I(t)=I$, 
the membrane potential of the LIF neuron $z_i$ changes over time is shown in Fig.~\ref{figure: 1b} (blue curve), which is computed according to Eq.~(\ref{LIF dynamics}).
Increasing or decreasing $w$, as shown by the $w+$ and $w-$ curves, will stretch the $v=f(t)$ curve in the $V$ direction. On the contrary, increasing or decreasing $\tau$ will stretch the $v=f(t)$ curve in the $t$ direction, and will not change the steady-state voltage of the neuron $z_i$ as $V(+\infty)=wI$.
Fig.~\ref{figure: effect of tau} illustrates the response of the neuron $z_i$ to instant input spikes at time $t = \{5, 80, 85, 90\}$ms, namely, $X(t)=w (\delta(t-5)+\delta(t-80)+\delta(t-85)+\delta(t-90))$ \footnote{$\delta(t)$ represents Dirac delta function. If $x \ne 0$, then $\delta(t)=0$. $\int_{-\infty}^{ \infty } \delta(t)~dt=1$.}. 
The neuron's response to instant input spike at $t=5$ indicates that a smaller $\tau$ (the $\tau-$ curve) leads to faster charge to the steady-state voltage and faster decay to the resting value, making the LIF neuron more sensitive to an instant spike. This sensitivity helps the neuron to capture instant variety in the input. 
In contrast, a smaller $w$ (the $w-$ curve) leads to a slower charge to the steady-state voltage without affecting decaying speed.
When there are three successive input spikes, the membrane potential of the neuron with a smaller $\tau$ (the $\tau-$ curve) will reach a higher value at a faster rate, which makes it easier to fire. 

To some extent, the effect of decreasing $\tau$ is similar to that of increasing $w$. Nevertheless, adjusting both $\tau$ and $w$ can bring some superior additional benefits. 
As mentioned above, changing both $\tau$ and $w$ can stretch the $v=f(t)$ curve, namely the neuron's response to a given input, in both $t$ direction and $V$ direction, which endows the neuron better fitting ability.

\subsection{Parametric Leaky Integrate-and-Fire model}
\label{Parametric Leaky Integrate-and-Fire Neuron}


We propose the Parametric Leaky Integrate-and-Fire (PLIF) spiking neuron model to learn both the synaptic weights and the membrane time constants of SNNs. The dynamics of the PLIF neuron can be described by Eq.~(\ref{LIF dynamics}). 

The SNNs with PLIF neurons follow the three rules:

(1). The membrane time constant $\tau$ is optimized automatically during training, rather than being set as a hyper-parameter manually before training.

(2). The membrane time constant $\tau$ is shared within the neurons in the same layer in SNNs, which is biologically plausible as the neighboring neurons have similar properties.

(3). The membrane time constant $\tau$ of neurons in different layers are distinct, making diverse phase-frequency responsiveness of neurons.

In fact, the proposed rules are able to increase the heterogeneity of neurons and the expressiveness of the resulting SNNs while effectively controlling computation costs.

\begin{figure}[t]
	\centering
	\includegraphics[width=0.4\textwidth, trim=0 350 600 0, clip]{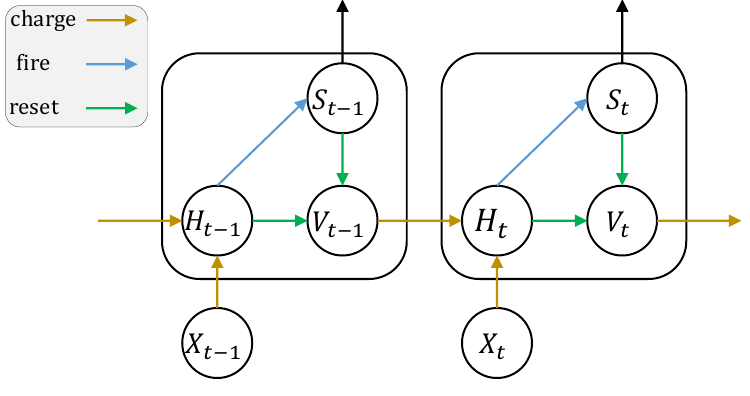}
	\caption{The general discrete spiking neuron model.}
	\label{fig:neuron model}
	\vspace{-0.5cm}
\end{figure}

For numerical simulations of PLIF neurons in SNNs, we need to consider a version of the parameters dynamics that is discrete in time. Specifically, by including the threshold-triggered firing mechanism and the reset of the membrane potential after firing, we can describe the dynamics of all kinds of spiking neurons with the following equations: 
\begin{align}
	H_{t} &= f(V_{t - 1}, X_{t}), \label{neural dynamics}\\
	S_{t} &= \Theta(H_{t} - V_{th}), \label{neural spiking}\\
	V_{t} &= H_{t}~(1 - S_{t}) + V_{reset}~S_{t}. \label{neural reset}
\end{align}
To avoid confusion, we use $H_{t}$ and $V_{t}$ to represent the membrane potential after neuronal dynamics and after the trigger of a spike at time-step $t$, respectively. $X_{t}$ denotes the external input, and $V_{th}$ denotes the firing threshold. $S_{t}$ denotes the output spike at time $t$, which equals 1 if there is a spike and 0 otherwise. Eq.~(\ref{neural spiking}) describes the spike generative process, where $\Theta(x)$ is the Heaviside step function and is defined by $\Theta(x) = 1$ for $x \ge 0$ and $\Theta(x) = 0$ for $x < 0$. 
Eq.~(\ref{neural reset}) illustrates that the membrane potential returns to $V_{reset}$ after eliciting a spike, which is called \textit{hard reset} and widely used in deep SNNs \cite{ledinauskas2020training}.

As shown in Fig.~\ref{fig:neuron model}, Eqs.~(\ref{neural dynamics}) - (\ref{neural reset}) build a general model to describe the discrete spiking neuron's action: charging, firing, and resetting. Specifically, Eq.~(\ref{neural dynamics}) describes the neuronal dynamics, and different spiking neuron models have different functions $f(\cdot)$. For example, the function $f(\cdot)$ for the LIF neuron and PLIF neuron is
\begin{align}
	H_{t} = V_{t-1} + \frac{1}{\tau}(-(V_{t-1} - V_{reset}) + X_{t}).
	\label{LIF iterative expression}
\end{align}


For PLIF neurons, directly optimizing the membrane time constant $\tau$ in Eq.~(\ref{LIF iterative expression}) may induce numerical instability as $\tau$ is in the denominator. Besides, Eq.~(\ref{LIF iterative expression}), as the discrete version of Eq.~(\ref{LIF dynamics}), is a valid approximation only when the time-step $\mathrm{d}t$ is smaller than $\tau$, that is, $\tau > 1$, which is ignored by \cite{rathi2020dietsnn, yin2020effective}. To avoid the above problems, we reformulate Eq.~(\ref{LIF iterative expression}) to the following equation with a trainable parameter $a$:
\begin{align}
	H_{t} = V_{t-1} + k(a)(-(V_{t-1} - V_{reset}) + X_{t}).
	\label{PLIF iterative expression}
\end{align}
Here $k(a)$ denotes the clamp function and $k(a) \in (0, 1)$, which ensures that $\tau = \frac{1}{k(a)} \in (1, +\infty)$.
In our experiments, $k(a)$ is the sigmoid activation
function, that is, $k(a) = \frac{1}{1+\mathrm{exp}(-a)}$.

\begin{figure}[t]
	\begin{center}
		\includegraphics[width=0.45 \textwidth, trim=4 98 180 0, clip]{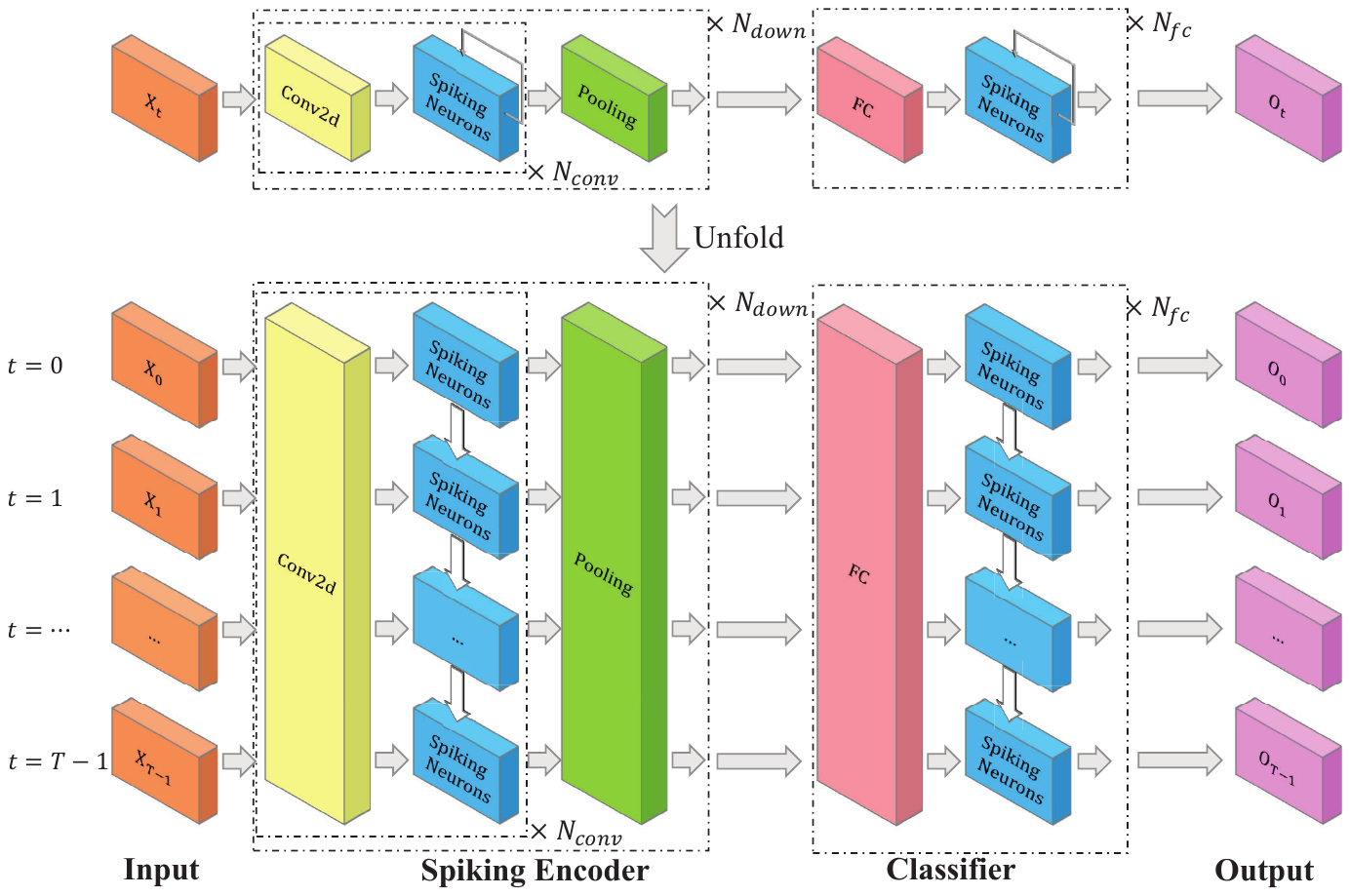}
		\caption{The general formulation of our networks and its unfolded formulation. $\times N_{conv}$ indicates there are $N_{conv}$ \textit{\{Conv2d-Spiking Neurons\}} connected sequentially. $\times N_{down}$ and $\times N_{fc}$ have the same meaning. Note that the network's parameters are shared at all time-steps.} 
		\label{fig:general network}
		\vspace{-0.5cm}
	\end{center}
\end{figure}

\subsection{RNN-like Expression of LIF and PLIF}
The LIF and PLIF neurons have a similar function as recurrent neural networks. Specifically, when $V_{reset}=0$, the neuronal dynamics of the LIF neuron and PLIF neuron (Eq.~\eqref{LIF iterative expression}) can be written as:
\begin{equation}
	H_{t} = \left(1 - \frac{1}{\tau}\right)V_{t-1} + \frac{1}{\tau}X_{t},
	\label{LIF iterative expression 2}
\end{equation}
where the integration progress $\frac{1}{\tau}X_{t}$ makes the LIF and PLIF neurons able to remember current input information, while the leakage progress $(1 - \frac{1}{\tau})V_{t-1}$ can be seen as forgetting some information from the past. Eq.~(\ref{LIF iterative expression 2}) shows that the balance between remembrance and forgetting is controlled by the membrane time constant $\tau$, which plays an analogous role as the gates in Long Short-Term Memory (LSTM) networks \cite{hochreiter1997long}.

\subsection{Network Formulation}
\label{networksf}
We propose a general formulation to build SNNs in this paper, which is illustrated in Fig.~\ref{fig:general network}. The SNN includes a spiking encoder network and a classifier network. The spiking encoder network consists of $N_{down}$ down-sample modules, each of which contains $N_{conv}$ repeated \textit{\{Conv2d-Spiking Neurons\}} and a pooling layer. The spiking encoder can extract features from inputs and convert them into the firing spikes at different time-steps. The classifier network consists of $N_{fc}$ repeated \textit{\{FC-Spiking Neurons\}}. Here \textit{Conv2d} denotes the 2D convolutional layer and \textit{FC} denotes the fully connected layer. Many previous works \cite{diehl2015unsupervised,lee2020enabling,shrestha2018slayer, ST-RSBP, LISNN, Han_2020_CVPR} used a Poisson encoder to convert images to spikes as input, while \cite{Bodo2017Conversion} suggested that this encoding introduces variability into the firing of the network and impairs its performance. Similar to \cite{Bodo2017Conversion, wu2019direct, rathi2020dietsnn}, the input is directly fed to our network without being firstly converted to spikes. In this situation, the image-spike encoding is done by the first \textit{\{Conv2d-Spiking Neurons\}} module, which can be seen as a learnable encoder. Note that synaptic connections, including convolutional layers and fully connected layers, are stateless, while the spiking neuron layers have self-connections in the temporal domain, as the unfolded network formulation shown in Fig.~\ref{fig:general network}. All parameters are shared at all time-steps.

\subsection{Spike Max-Pooling}
\label{sectionpooling}
\begin{figure}[t]
	\centering
	\subfigure[Spike max-pooling]{\includegraphics[width=0.42\textwidth, trim=30 100 20 100, clip]{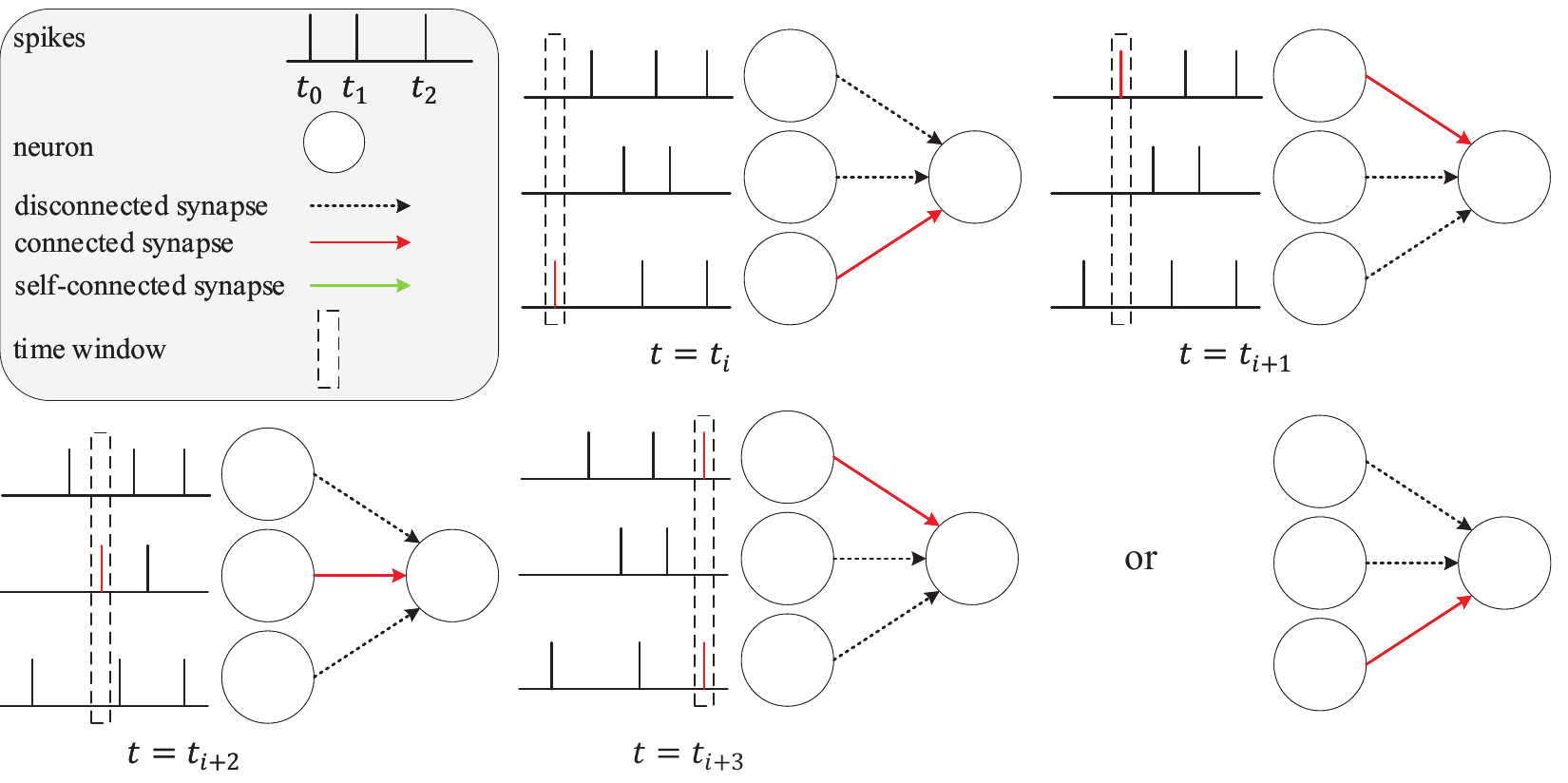}}
	\subfigure[Unfolded computation graph]{\includegraphics[width=0.42\textwidth, trim=30 240 20 100, clip]{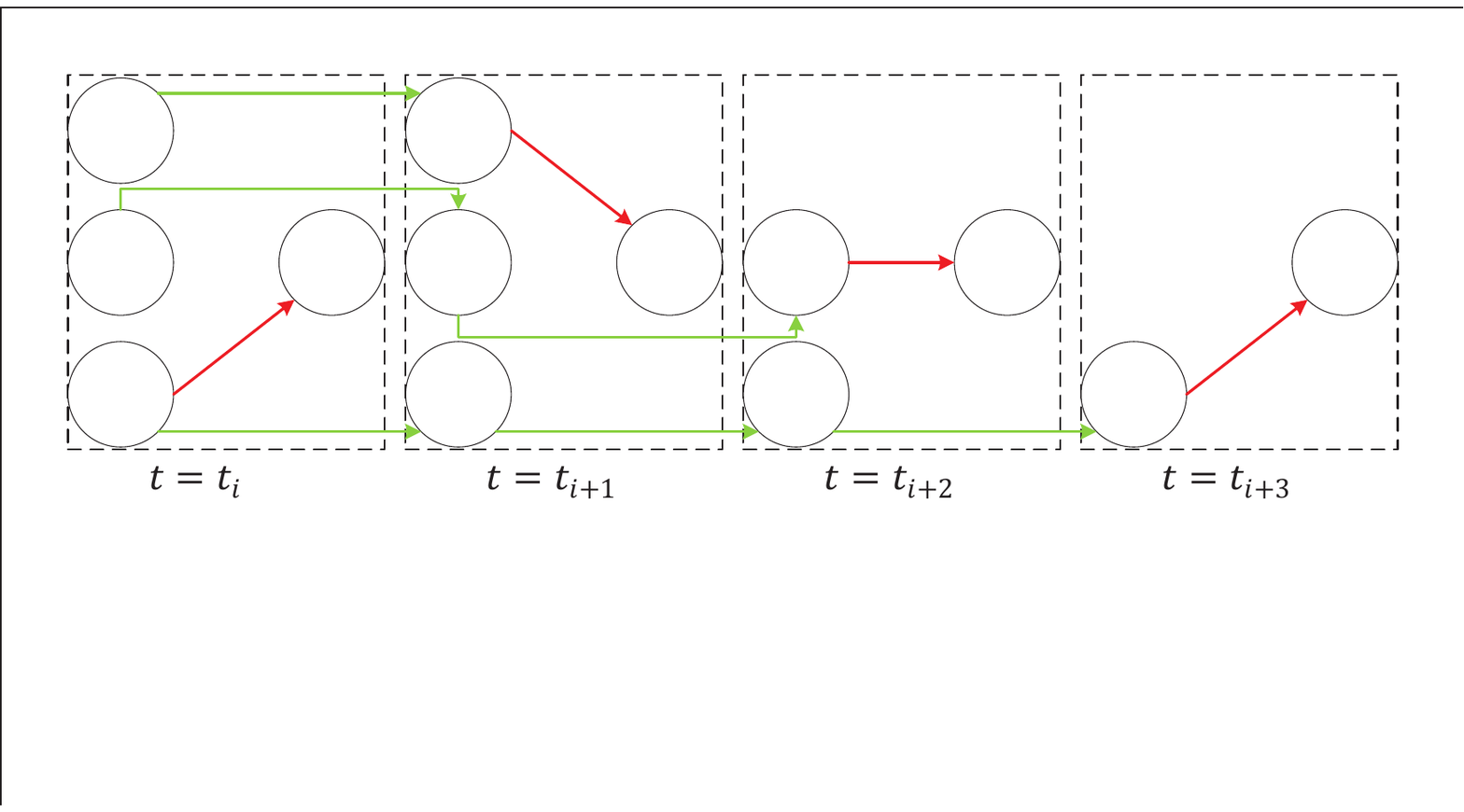}}
	\caption{Spike max-pooling regulates connections dynamically. (a) An example of three presynaptic neurons and one postsynaptic neuron with spike max-pooling. At every time-step, only the neuron that fires a spike can connect to the postsynaptic neuron. When more than one neuron fire at the same time-step, the neuron that can connect to the postsynaptic neuron is randomly selected. (b) The unfolded computation graph of (a).}
	\label{fig:max pooling}
\end{figure}
The pooling layer is widely used to reduce the size of feature maps and to extract compact representation in convolutional ANNs, as well as SNNs.
Most previous studies \cite{sengupta2019going, LISNN, Rathi2020Enabling} preferred to use the average-pooling in SNNs as they found that max-pooling in SNNs leads to significant information loss. 
We argue that the max-pooling is consistent with the SNNs' temporal information processing ability and can increase SNNs' fitting capability in temporal tasks and reduce the computation cost for the next layer.

Specifically, the max-pooling layers are behind spiking neuron layers in our model (Fig.~\ref{fig:general network}), and the max-pooling operation is carried on spikes. Different from all neurons that transmit information to the next layer equally in the average-pooling window, only the neuron that fires a spike in the max-pooling window can transmit information to the next layer. 
Therefore, the max-pooling layer introduces the winner-take-all mechanism, allowing the fired neuron to communicate with the next layer and ignoring other neurons in the pooling window. Another attractive property is that the max-pooling layer will regulate connections dynamically (Fig.~\ref{fig:max pooling}). The spiking neuron's membrane potential $V_{t}$ will return to $V_{reset}$ after firing a spike. It is hard for a spiking neuron to fire again as recharging needs time. However, if the neurons in the max-pooling window fire asynchronously, they will be connected to the postsynaptic neuron in turn, which makes the postsynaptic neuron resembles to connect a continuously firing presynaptic neuron and easier to fire. The winner-take-all mechanism in the spatial domain and time-variant topology in the temporal domain achieved by max-pooling can increase SNNs' fitting capability in temporal tasks, such as classifying the CIFAR10-DVS dataset. It is worth noting that the outputs of the max-pooling layer are still binary, while the outputs of the average-pooling layer are float. The matrix multiplication and element-wise multiplication operation on spikes can be accelerated by replacing \textit{multiplication} $*$ with \textit{logical AND} $\&$, which is also the advantage of SNNs compared with ANNs.

\subsection{Training Framework}
\label{sectionlearning}
Here we combine the neuron model (Fig.~\ref{fig:neuron model}) and network formulation (Fig.~\ref{fig:general network}) to drive the backpropagation training algorithm for SNNs. Denote the simulating time-steps as $T$ and classes number as $C$, the output $\boldsymbol O = [ o_{t, i} ] $ is a $C \times T$ tensor. For a given input with label $l$, we encourage the neuron that represents class $l$ to have the highest excitatory level while other neurons should remain silent. So the target output is defined by $\boldsymbol Y=[y_{t, i}]$ with $y_{t, i} = 1$ for $i=l$ and $y_{t, i} = 0$ for $i \neq l$. The loss function is defined by the mean squared error (MSE) $L = MSE(\boldsymbol O, \boldsymbol Y) = \frac{1}{T} \sum^{T-1}_{t = 0} L_{t} = \frac{1}{T} \sum^{T-1}_{t = 0} \frac{1}{C} \sum^{C-1}_{i = 0} (o_{t, i} - y_{t,i})^2$. And the predicted label $l_{p}$ is regarded as the index of the neuron with the maximum firing rate $l_{p} = \mathop{\arg\max_{i}} \frac{1}{T}\sum_{t=0}^{T-1} o_{t, i}$.

Here we suppose that $a^{i}$ represents the learnable parameter of the PLIF neurons in the $i$-th layer in the network. At time-step $t$, the vectors $\boldsymbol H_{t}^{i}$ and $\boldsymbol V_{t}^{i}$ represent the membrane potential after neuronal dynamics and after reset, the vector $\boldsymbol V_{th}^{i}$ and $\boldsymbol V_{reset}^{i}$ represents the threshold and reset potential, respectively. The weighted inputs from the previous layer are $\boldsymbol X_{t}^{i} = \boldsymbol W^{i-1} \boldsymbol I_{t}^{i}$. $\boldsymbol S_{t}^{i} = [s_{t,j}^{i}]$ denotes the output spike at time-step $t$, where $s_{t,j}^{i} = 1$ if the $j$-th neuron fires a spike, else $s_{t,j}^{i} = 0$. The gradients backward from the next layer are $\frac{\partial L_{t}}{\partial \boldsymbol S_{t}^{i}}$. According to Fig.~\ref{fig:neuron model} and Fig.~\ref{fig:general network}, we can calculate the gradients recursively:
\begin{align}
	& \frac{\partial L}{\partial \boldsymbol H_{t}^{i}} = \frac{\partial L}{\partial \boldsymbol H_{t+1}^{i}} \frac{\partial \boldsymbol H_{t+1}^{i}}{\partial \boldsymbol H_{t}^{i}} + \frac{\partial L_{t}}{\partial \boldsymbol H_{t}^{i}}\\
	& \frac{\partial \boldsymbol H_{t+1}^{i}}{\partial \boldsymbol H_{t}^{i}} = \frac{\partial \boldsymbol H_{t+1}^{i}}{\partial \boldsymbol V_{t}^{i}} \frac{\partial \boldsymbol V_{t}^{i}}{\partial \boldsymbol H_{t}^{i}} \\
	& \frac{\partial L_{t}}{\partial \boldsymbol H_{t}^{i}} = \frac{\partial L_{t}}{\partial \boldsymbol S_{t}^{i}} \frac{\partial \boldsymbol S_{t}^{i}}{\partial \boldsymbol H_{t}^{i}}
\end{align}
According to Eq.~(\ref{PLIF iterative expression}), Eq.~(\ref{neural spiking}), and Eq.~(\ref{neural reset}) we can get
\begin{align}
	& \frac{\partial \boldsymbol H_{t+1}^{i}}{\partial \boldsymbol V_{t}^{i}} = 1 -k(a^{i}) \\
	& \frac{\partial \boldsymbol V_{t}^{i}}{\partial \boldsymbol H_{t}^{i}} = 1 - \boldsymbol S_{t} + (\boldsymbol V_{reset}^{i} - \boldsymbol H_{t}^{i}) \frac{\partial \boldsymbol S_{t}^{i}}{\partial \boldsymbol H_{t}^{i}} \\
	& \frac{\partial \boldsymbol S_{t}^{i}}{\partial \boldsymbol H_{t}^{i}} = \Theta'(\boldsymbol H_{t}^{i} - \boldsymbol V_{th}^{i}) \\
	& \frac{\partial \boldsymbol H_{t}^{i}}{\partial \boldsymbol X_{t}^{i}} = k(a^{i}) \\
	\begin{split}
		\frac{\partial \boldsymbol H_{t}^{i}}{\partial a^{i}} = (-(\boldsymbol V_{t-1}^{i} - \boldsymbol V_{reset}^{i}) + \boldsymbol X_{t}^{i}) k'(a^{i}) \\+ \frac{\partial \boldsymbol H_{t}^{i}}{\partial \boldsymbol V_{t-1}^{i}} \frac{\partial \boldsymbol V_{t-1}^{i}}{\partial \boldsymbol H_{t-1}^{i}} \frac{\partial \boldsymbol H_{t-1}^{i}}{\partial a^{i}}
	\end{split}
\end{align}

Finally, we can get the gradients of the learnable parameters:
\begin{align}
	&\frac{\partial L}{\partial a^{i}} = \sum_{t=0}^{T-1}\frac{\partial L}{\partial \boldsymbol H_{t}^{i}} \frac{\partial \boldsymbol H_{t}^{i}}{\partial a^{i}} \label{eq gd1}\\
	&\frac{\partial L}{\partial \boldsymbol W^{i-1}} = \sum_{t=0}^{T-1}\frac{\partial L}{\partial \boldsymbol H_{t}^{i}} \frac{\partial \boldsymbol H_{t}^{i}}{\partial \boldsymbol X_{t}^{i}}\boldsymbol I_{t}^{i} \label{eq gd2}
\end{align}
Note that $\frac{\partial *}{\partial \boldsymbol S_{t}^{i}} = 0$ when $t \geq T$, $\boldsymbol V_{-1}^{i} = \boldsymbol V_{reset}^{i}$. We use derivative of the surrogate function $\sigma (x)$ to define the derivative of spiking function $\Theta(x)$ (see supplementary). $k(x)$ is the clamp function.

\section{Experiments}
We evaluate the performance of SNNs with PLIF neurons and spike max-pooling for classification tasks on both traditional static MNIST, Fashion-MNIST, CIFAR-10 datasets, and neuromorphic N-MNIST, CIFAR10-DVS, and DVS128 Gesture datasets. More details of the training can be found in the supplementary.

\begin{table}
	\centering
	\begin{tabular}{cccc}
		\hline
		\textbf{Dataset} & $N_{conv}$ & $N_{down}$ & $N_{fc}$ \\
		\hline
		*MNIST &1 &2 &2 \\
		CIFAR-10 &3 &2 &2 \\
		CIFAR10-DVS &1 &4 &2 \\
		DVS128 Gesture &1 &5 &2 \\
		\hline
	\end{tabular}
	\caption{Network structures for different datasets. $N_{conv}$, $N_{down}$ and $N_{fc}$ are defined in Fig.~\ref{fig:general network}. \textit{*MNIST} denotes MNIST, Fashion-MNIST and N-MNIST datasets.}
	\label{tab:network structures}
\end{table}

\subsection{Network Structure} 
\label{secthion: Network Structure}
The network structures of SNNs for different datasets are shown in Tab.~\ref{tab:network structures}. We set $kernel~size = 3$, $stride = 1$ and $padding = 1$ for all \textit{Conv2d} layers. The $out~channels$ of \textit{Conv2d} layers is 256 for CIFAR-10 dataset and 128 for all other datasets. A batch normalization (\textit{BN}) layer is added after each \textit{Conv2d} layer. As the parameters of a \textit{BN} layer can be absorbed in its front \textit{Conv2d} layer \cite{Bodo2017Conversion}, we can remove \textit{BN} in the SNNs for inference. All pooling layers set $kernel~size = 2$ and $stride = 2$. For all networks, the $out~features$ of the first $FC$ layer is a quarter of the $in~features$, and the $out~features$ of the second $FC$ layer is $M\cdot C$, where $C$ is the classes number and $M$ is the neurons of a population to represent one class. A dropout layer \cite{lee2020enabling} is placed before each $FC$ layer. A voting layer after the output spiking neurons layer is used to boost classifying robustness. The voting layer is implemented by average-pooling with $kernel~size = M$ and $stride = M$. We set $M=10$ for all datasets. We use the average-pooling to implement democratic voting, such that the minority is subordinate to the majority. Using max-pooling to vote may result in a dictatorship, as the minority will not be involved in the computation graph (see Fig.~\ref{fig:max pooling}) and using $M$ neurons to represent one class will degenerate into using one neuron. 

\begin{table*}
	\centering
	\scalebox{0.95}
	{
		\begin{tabular}{cccccccc}
			\toprule
			\textbf{Model} & \textbf{Method} & \textbf{\makecell[l]{Accuracy \\ MNIST}} & \textbf{\makecell[l]{Accuracy \\ Fashion-MNIST}} & \textbf{\makecell[l]{Accuracy \\ CIFAR-10}} & \textbf{\makecell[l]{Accuracy \\ N-MNIST}} &\textbf{\makecell[l]{Accuracy \\ CIFAR10-DVS}} &\textbf{\makecell[l]{Accuracy \\ DVS128 Gesture}}\\
			\midrule
			
			\cite{hunsberger2015spiking} & ANN2SNN & 98.37\% & - & 82.95\% & - & - & - \\
			\cite{Bodo2017Conversion} & ANN2SNN & 99.44\% & - & 88.82\% & - & - & - \\
			\cite{sengupta2019going} & ANN2SNN & - & - & 91.55\% & - & - & - \\
			
			\cite{Han_2020_CVPR} & ANN2SNN & - & - & \textbf{93.63\%} & - & - & -  \\
			
			\cite{lee2016training} & Spike-based BP & 99.31\% & - & - & 98.74\% & - & - \\ 

			\cite{wu2018STBP} & Spike-based BP & 99.42\% & - & - & 98.78\% & 50.7\% & - \\
			\cite{shrestha2018slayer} & Spike-based BP & 99.36\% & - & - & 99.2\% & - & 93.64\% \\
			
			\cite{10.3389/fnins.2020.00424} & Spike-based BP & - & - & - & 96\% & - & \textbf{95.54\%} \\
			\cite{HM-2BP} & Spike-based BP & 99.49\% & - & - & 98.84\% & - & - \\
			
			\cite{ST-RSBP} & Spike-based BP & \textbf{99.62\%} & 90.13\% & - & - & - & - \\
			
			\cite{wu2019direct} & Spike-based BP & - & - & 90.53\% & \textbf{99.53\%} & \textbf{60.5\%} & - \\
			
			\cite{lee2020enabling} & Spike-based BP & 99.59\% & - & 90.95\% & 99.09\% & - & - \\
			\cite{LISNN} & Spike-based BP & 99.5\% & \textbf{92.07\%} & - & 99.45\% & - & - \\
			
			\cite{liu2020effective} & Spike-based BP & - & - & - & 96.3\% & 32.2\% & - \\
			
			\cite{xing2020new} & Spike-based BP & - & - & - & - & - & 92.01\% \\

			\cite{ijcai2020-0388} & Spike-based BP & 99.46\% & - & - & 99.39\% & - & \makecell{96.09\% \\(10 classes)} \\
			
			\cite{HE2020108} & Spike-based BP & - & - & - & 98.28\% & - & 93.40\% \\
			
			
			
			\cite{rathi2020dietsnn} & \makecell{ANN2SNN and\\ Spike-based BP} & - & - & 92.64\% & - & - & - \\
			
			\cite{sironi2018hats} & HATS & - & - & - & 99.1\% & 52.4\% & - \\
			
			\cite{bi2019graph-based} & GCN & - & - & - & 99.0\% & 54.0\% & - \\
			
			\midrule
			Ours & Spike-based BP & 99.72\% & 94.38\% & 93.50\% & 99.61\% & 74.80\% & 97.57\%\\
			%
			\bottomrule
		\end{tabular}
	}
	\caption{Performance comparison between the proposed method and the state-of-the-art methods on different datasets. The highest accuracies of previous works are in bold.}
	\label{tab:ACC CMP}
	
\end{table*}
\subsection{Comparison with the State-of-the-Art}

Tab.~\ref{tab:ACC CMP} shows the accuracies of the proposed methods (PLIF neurons with $\tau_{0}=2$, max-pooling)
and other comparing methods on both traditional static MNIST, Fashion-MNIST, CIFAR-10 datasets, and neuromorphic N-MNIST, CIFAR10-DVS, DVS128 Gesture datasets. We set the same training hyperparameters for all datasets (see supplementary). As shown in Tab.~\ref{tab:ACC CMP}, we achieve the highest accuracies on all datasets except for CIFAR-10. The accuracy on CIFAR-10 is slightly lower than \cite{Han_2020_CVPR}, which is based on ANN2SNN conversion. However, they only applied to static images as ANN2SNN is ill-suited to neuromorphic datasets. Different from them, our method is also applicable to neuromorphic datasets and outperforms the spike-based BP SOTA accuracy. 



\begin{table}
	\centering
	\scalebox{0.975}
	{
		\begin{tabular}{lp{1.0cm}p{1.6cm}p{1.8cm}}
			\hline
			\textbf{Dataset} & \textbf{SOTA} & \textbf{SOTA's} $T$ & \textbf{ours} $T$ \\
			\hline
			MNIST & \cite{ST-RSBP} & 400 & 8\\
			Fashion-MNIST & \cite{LISNN} & 20 & 8\\
			CIFAR-10 & \cite{Han_2020_CVPR} & 2048 & 8\\
			N-MNIST & \cite{wu2019direct} & 59-64 & 10\\
			CIFAR10-DVS & \cite{wu2019direct} & 230-292 & 20\\
			DVS128 Gesture & \cite{10.3389/fnins.2020.00424} &\makecell[l]{500(training) \\ 1800(testing)} & 20\\
			\hline
		\end{tabular}
	}
	\caption{The time-steps of previous SOTA works and ours on each dataset.}
	
	\label{tab:previous stoa T}
\end{table}

Tab.~\ref{tab:previous stoa T} compares the number of time-steps of our method and the previous works that achieve the best performance on each dataset. It can be found that the proposed method takes fewer time-steps than all the other methods. For example, our method uses up to $256\times$ fewer inference time-steps compared to ANN2SNN conversion \cite{Han_2020_CVPR}. Thus our method can not only decrease the memory consumption and the training time but also increase inference speed greatly.

\subsection{Ablation Study} \label{Ablation Study}
\begin{figure}
	\centering
	\subfigure[Fashion-MNIST]{\includegraphics[width=0.22\textwidth,trim=0 0 0 0,clip]{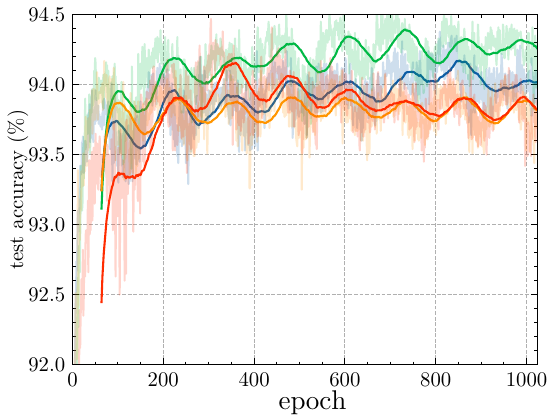}}
	\subfigure[CIFAR-10]{\includegraphics[width=0.22\textwidth,trim=0 2 0 0,clip]{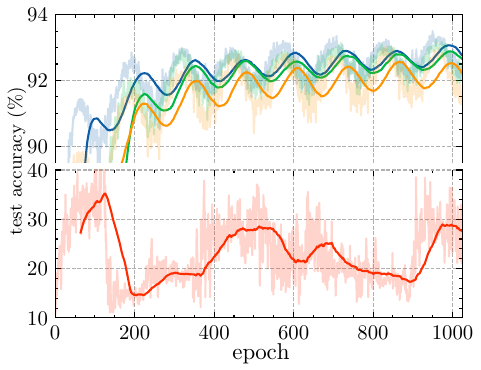}}
	\subfigure{\includegraphics[width=0.44\textwidth,trim=2 180 2 0,clip]{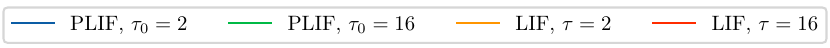}}
	\addtocounter{subfigure}{-1}
	\subfigure[CIFAR10-DVS]{\includegraphics[width=0.22\textwidth,trim=0 0 0 0,clip]{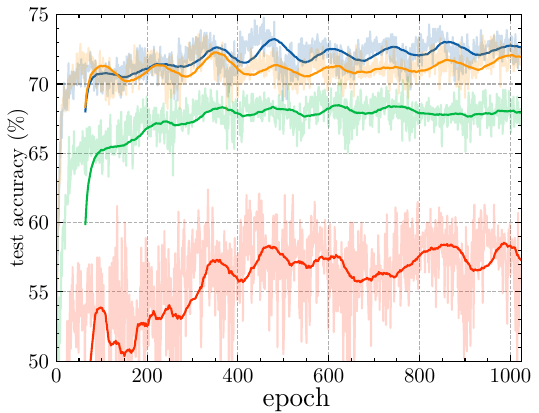}}
	\subfigure[DVS128 Gesture]{\includegraphics[width=0.22\textwidth,trim=0 0 0 0,clip]{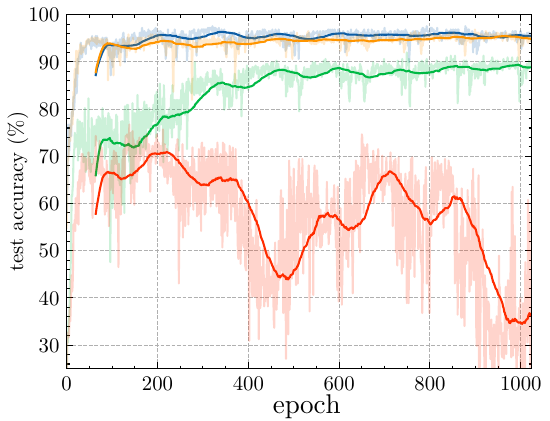}}
	\caption{The test accuracy of PLIF vs. LIF neurons on different datasets during training. The shaded curves indicate the origin data. The solid curves are 64-epoch moving averages.}
	\label{figure: PLIF v.s. LIF}
\end{figure}

We conduct extensive ablation studies to evaluate PLIF neurons and max-pooling on four challenging datasets. We first study the effect of PLIF neurons. In this experiment, we train the same SNNs with PLIF neurons and LIF neurons respectively, and compare the test accuracy.
As shown in Tab.~\ref{table: LIF vs PLIF}, if the initial membrane time constant $\tau_{0}$ of PLIF neurons is set equal to the membrane time constant $\tau$ of LIF neurons, the test accuracy of the SNNs with PLIF neurons is always higher than that with LIF neurons. This is due to the membrane time constants of PLIF neurons in different layers can be different after learning, which better represents the heterogeneity of neurons. 
Fig.~\ref{figure: PLIF v.s. LIF} illustrates the test accuracy of PLIF vs. LIF neurons during training. As can be seen, the accuracy and convergence speed of the SNNs with LIF neurons decrease seriously if the initial value of the membrane time constant is not reasonable (red curve). In contrast, the PLIF neurons can learn the appropriate membrane time constants and achieve better performance (green curve). 
\begin{table}
	\centering
	\scalebox{0.635}
	{
		\begin{tabular}{ccccc}
			\toprule
			\textbf{Neuron} & \textbf{Fashion-MNIST} & \textbf{CIFAR-10} &\textbf{CIFAR10-DVS} &\textbf{DVS128 Gesture}\\
			\midrule
			PLIF($\tau_{0}=2$) & \textbf{94.38\%} & \textbf{93.50\%} & \textbf{74.80\%} & \textbf{97.57\%}\\
			
			LIF($\tau=2$) & 94.17\% & 93.03\% & 73.60\% & 96.88\%\\
			\midrule
			PLIF($\tau_{0}=16$) & \textbf{94.65\%} & \textbf{93.23\%} & \textbf{70.50\%} & \textbf{92.01\%}\\
			
			LIF($\tau=16$) & 94.47\% & 47.50\% & 62.40\% & 76.74\%\\
			\bottomrule
		\end{tabular}
	}
	\caption{Accuracy of using PLIF/LIF.}
	\label{table: LIF vs PLIF}
\end{table}

\begin{figure}
	\centering
	\subfigure[The change of $\tau(i)$ during training on CIFAR-10.]{\includegraphics[width=0.5\textwidth,trim=0 0 0 0,clip]{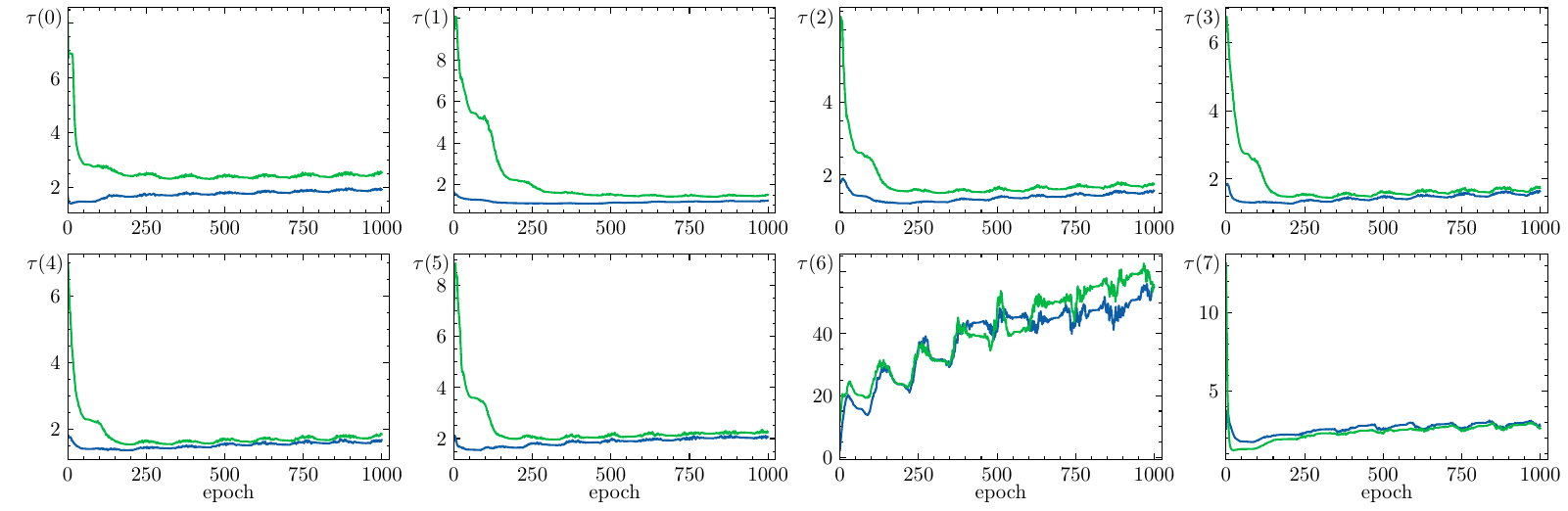}\label{figure: tau during training-a}}
	\subfigure{\includegraphics[width=0.25\textwidth,trim=0 164 40 0,clip]{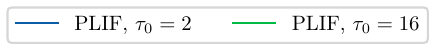}}
	\addtocounter{subfigure}{-1}
	\subfigure[The change of $\tau(i)$ during training on CIFAR10-DVS.]{\includegraphics[width=0.5\textwidth,trim=0 0 0 0,clip]{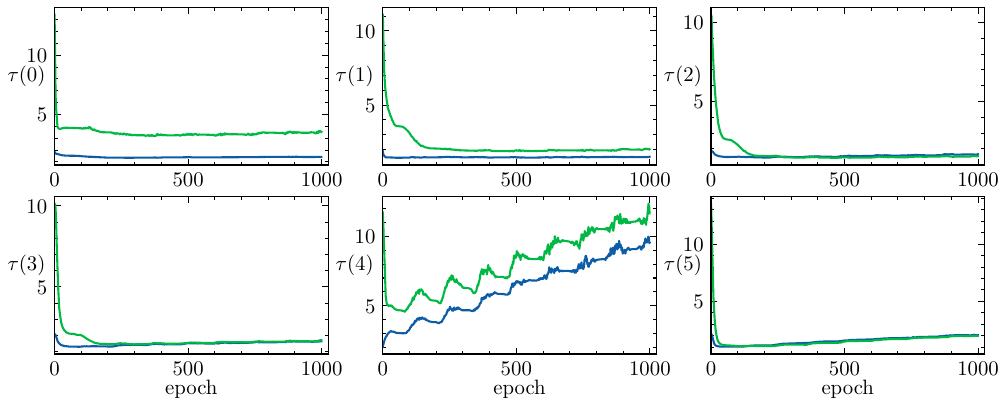}\label{figure: tau during training-b}}
	\caption{The change of membrane time constants in different layers during training with different initial values. $\tau(i)$ represents the  membrane time constant $\tau$ of the $i$-th PLIF neurons layer.}
	\label{figure: tau during training}
	\vspace{-0.1cm}
\end{figure}

To analyze the influence of initial values in PLIF neurons, we show how the membrane time constants of the neurons in each layer change during learning with respect to different initial values. As shown in Fig.~\ref{figure: tau during training}, the membrane time constants with different initial values in each layer tend to gather during training, which indicates that the PLIF neurons are robust to initial values. Note that $\tau(6)$ in Fig.~\ref{figure: tau during training-a}\ and $\tau(4)$ in Fig.~\ref{figure: tau during training-b} tend to infinity. This could be explained as follows.
The PLIF neurons with the membrane time constants $\tau(4)$ and $\tau(6)$ in two SNNs are behind the first FC layer with weight $W_{fc}$. We check the training logs and find that the distribution, mean and variance of $\frac{W_{fc}}{\tau} (\tau=\tau(4)~\text{or}~\tau(6))$ converge after dozens of epochs (see supplementary). Refer to the dynamics of PLIF neurons (Eq.~(\ref{LIF iterative expression})) with $X_{t}=W_{fc}I_{t}$ and $\frac{1}{\tau} \rightarrow 0$, we can find $H_{t} \rightarrow V_{t-1} + \frac{W_{fc}}{\tau}I_{t}$. It means that the PLIF neurons after the first FC layer are learning to become the Non-Leaky-Integrate-and-Fire neurons. 

We further study the effect of max-pooling. Tab.~\ref{table: max vs avg} compares the accuracy of the proposed SNNs with max-pooling/average-pooling on four challenging datasets. The performance of max-pooling is similar to that of average-pooling, which indicates that the previous conclusion that max-pooling results in significant information loss in SNNs is not reasonable. Remarkably, the max-pooling gets slightly higher accuracies on CIFAR-10, CIFAR10-DVS, and DVS128 Gesture datasets, showing its better fitting capability in complex tasks.
\begin{table}
	\centering
	\scalebox{0.685}
	{
		\begin{tabular}{ccccc}
			\toprule
			\textbf{Pooling} & \textbf{Fashion-MNIST} & \textbf{CIFAR-10} &\textbf{CIFAR10-DVS} &\textbf{DVS128 Gesture}\\
			\midrule
			Average & \textbf{94.74\%} & 93.30\% & 72.70\% & 97.22\%\\
			
			Max & 94.38\% & \textbf{93.50\%} & \textbf{74.80\%} & \textbf{97.57\%}\\
			\bottomrule
		\end{tabular}
	}
	\caption{Accuracy of using max-pooling/average-pooling.}
	\label{table: max vs avg}
\end{table}

\section{Conclusion}
In this work, we proposed the Parametric Leaky Integrate-and-Fire (PLIF) neuron to incorporate the learnable membrane time parameter into SNNs. We show that the SNNs with the PLIF neurons outperform state-of-the-art comparing methods on both static and neuromorphic datasets. Besides, we show that the SNNs made of PLIF neurons are more robust to initial values and can learn faster than SNNs consist of LIF neurons.
We also reevaluate the performance of max-pooling and average-pooling in SNNs and find the previous works underestimate the performance of max-pooling. We recommend using max-pooling in SNNs for its lower computation cost, higher temporal fitting capability, and the characteristic to receive spikes and output spikes rather than floating values as average-pooling.

\section{Acknowledgment}
This work is supported by grants from the National Natural Science Foundation of China under contracts No.62027804, No.61825101, and No.62088102.

\begin{appendices}
\section{Supplementary Materials}
\section{Reproducibility}
All experiments are implemented by SpikingJelly \cite{SpikingJelly}. All of the source codes, training logs are available on GitHub. To maximize reproducibility, we use identical seeds in all codes.

\begin{table}[b]
	\centering
	\begin{tabular}{lp{5.2cm}p{0.8cm}}
		\hline
		\textbf{Dataset} & \textbf{Network Structure}\\
		\hline
		*MNIST & \makecell[l]{\{c128k3s1-BN-PLIF-MPk2s2\}*2-\\DP-FC2048-PLIF-DP-FC100-PLIF-\\APk10s10}\\
		
		CIFAR-10 & \{\{c256k3s1-BN-PLIF\}*3-MPk2s2\}*2-DP-FC2048-PLIF-DP-FC100-PLIF-APk10s10 \\
		
		CIFAR10-DVS &
		\{c128k3s1-BN-PLIF-MPk2s2\}*4-DP-FC512-PLIF-DP-FC100-PLIF-APk10s10 \\
		
		DVS128 Gesture &
		\{c128k3s1-BN-PLIF-MPk2s2\}*5-DP-FC512-PLIF-DP-FC110-PLIF-APk10s10 \\
		\hline
	\end{tabular}
	\caption{Detailed network structures for different datasets. \textit{*MNIST} represents MNIST, Fashion-MNIST, and N-MNIST datasets.}
	\label{tab:detailed network structures}
\end{table}

\section{Network Structure Details}
Tab.~\ref{tab:detailed network structures} illustrates the details of the network structures for different datasets. \textit{c128k3s1} represents the convolutional layer with $output~channels = 128$, $kernel~size = 3$ and $stride = 1$. \textit{BN} is the batch normalization. \textit{MPk2s2} is the max-pooling layer with $kernel~size = 2$ and $stride = 2$. \textit{PLIF} is the PLIF spiking neurons layer. \textit{DP} represents the dropout layer \cite{lee2020enabling}. \textit{FC2048} represents the fully connected layer with $output~features = 2048$. The symbol \textit{\{\}*} indicates the repeated structure. For example, \textit{\{c128k3s1-BN-PLIF-MPk2s2\}*2} means that there are two \textit{\{c128k3s1-BN-PLIF-MPk2s2\}} modules connected sequentially. The last layer \textit{APk10s10} is the voting layer, which is implemented by an average-pooling layer with $kernel~size = 10$ and $stride = 10$.

\section{Training Algorithm to Fit Target Output}
After defining the derivative of the spike generative process, the parameters of SNNs can be trained by gradient descent algorithms as that in ANNs. Classification, which is the task in this paper, as well as other tasks for both ANNs and SNNs, can be seen as optimizing parameters of the network to fit a target output when given a specific input. The gradient descent algorithm for SNNs to fit a target output is derived in the main text (Eq.(\ref{eq gd1}) and Eq.(\ref{eq gd2})), and is as follows:
\begin{algorithm}[H]
	\caption{Gradient Descent Algorithm for SNNs to Fit Target Output}
	\textbf{Require:} learning rate $\epsilon$, network's parameter $\boldsymbol \theta$, total simulating time-steps $T$, input $\boldsymbol X = \{\boldsymbol X_{0}, \boldsymbol X_{1}, ..., \boldsymbol X_{T-1}\}$, target output $\boldsymbol Y = \{\boldsymbol Y_{0}, \boldsymbol Y_{1}, ..., \boldsymbol Y_{T-1}\}$, loss function $L=\mathcal L(\boldsymbol O, \boldsymbol Y)$
	
	initialize $\boldsymbol \theta$
	
	create an empty list $\boldsymbol S = \{ \}$
	
	\algorithmicfor{ $t \gets 0, 1, ... T-1$}
	
	~~~input $\boldsymbol X_{t}$ to network, get output spikes $\boldsymbol S_{t}$
	
	~~~append $\boldsymbol S_{t}$ to $\boldsymbol S = \{\boldsymbol S_{0}, \boldsymbol S_{1}, ..., \boldsymbol S_{t-1}\}$
	
	calculate loss $L=\mathcal L(\boldsymbol Y, \boldsymbol O)$
	
	update parameter $\boldsymbol \theta =\boldsymbol \theta - \epsilon \cdot \nabla_{\boldsymbol \theta} L$
\end{algorithm}
Here the loss function $L=\mathcal L(\boldsymbol O, \boldsymbol Y)$ is a distance measurement between $\boldsymbol Y$ and $\boldsymbol S$, e.g., the mean squared error (MSE) in the main text.

\section{Introduction of the Datasets}
\paragraph{MNIST}
The MNIST dataset of handwritten digits comprises $28 \times 28$ gray-scale images which are labeled from 0 to 9. The MNIST dataset includes 60,000 training images and 10,000 test images.
\paragraph{Fashion-MNIST}
Similar to the MNIST dataset, the Fashion-MNIST dataset consists of a training set of 60,000 examples and a test set of 10,000 examples. Each example in the Fashion-MNIST dataset is a $28 \times 28$ gray-scale image with a label from 0 to 9.
\paragraph{CIFAR-10}
The CIFAR-10 dataset consists of 60,000 natural images in 10 classes, with 6,000 images per class. The number of the training images is 50,000, and that of the test images is 10,000.
\paragraph{N-MNIST}
The Neuromorphic-MNIST (N-MNIST) dataset is a spiking version of the MNIST dataset recorded by the neuromorphic sensor. It was converted from MNIST by mounting the ATIS sensor on a motorized pan-tilt unit and moving the sensor while recording MNIST examples on an LCD monitor. It consists of 60,000 training examples and 10,000 test examples.

\paragraph{CIFAR10-DVS}
The CIFAR10-DVS dataset is the neuromorphic version of the CIFAR-10 dataset. It is composed of 10,000 examples in 10 classes, with 1000 examples in each class. As the CIFAR10-DVS dataset does not separate data into training and testing sets, in each class, we choose the first 9000 samples for training and the rest 1000 samples for testing, which is similar to \cite{wu2019direct}.

\paragraph{DVS128 Gesture}
The DVS128 Gesture dataset is recorded by a DVS128 camera, which contains 11 kinds of hand gestures from 29 subjects under 3 kinds of illumination conditions.

\section{Preprocessing}
\noindent
\textbf{Static Datasets.}~
We apply data normalization on all static datasets to ensure that input images have zero mean and unit variance. Besides, random horizontal flipping and cropping on MNIST and CIFAR-10 are conducted to avoid over-fitting. We do not use these augmentations on Fashion-MNIST as images in this dataset are tidy.

\noindent
\textbf{Neuromorphic Datasets.}~The data in neuromorphic datasets usually take the form of address event representation (AER)  $E(x_{i}, y_{i}, t_{i}, p_{i})$ ($i=0,1,...,N-1$) to represent the event location in the asynchronous stream, the timestamp, and the polarity. As the number of events is large, e.g. more than one million in CIFAR10-DVS, we split the events into $T$ slices with nearly the same number of events in each slice and integrate events to frames. The new representation  $F(j, p, x, y)$ ( $0 \leq j \leq T-1$) is the summation of event data in the $j$-th slice:
\begin{align}
	F(j, p, x, y)  = \sum_{i = j_{l}}^{j_{r} - 1} \mathcal{I}_{p, x, y}(p_{i}, x_{i}, y_{i}),
\end{align}
where $\mathcal{I}_{p, x, y}(p_{i}, x_{i}, y_{i})$ is an indicator function and it equals 1 only when $(p, x, y) = (p_{i}, x_{i}, y_{i})$. $j_l$ and $j_r$ are the minimal and the maximal timestamp indexes in the $j$-th slice. $j_{l}  = \left\lfloor \frac{N}{T}\right \rfloor \cdot j$, $j_{r}=\left \lfloor \frac{N}{T} \right \rfloor \cdot (j + 1)$ if $j <  T - 1$ and $N$ if $j = T-1$. Here $\lfloor \cdot \rfloor$ is the floor operation.
Note that $T$ is also the number of time-steps in our experiments.


Similar event-to-frame integrating methods for pre-processing neuromorphic datasets are widely used in both ANNs~\cite{HE2020108, neil2016phased, 7539039} and SNNs~\cite{wu2019direct, wu2020brain, LISNN, xing2020new, HE2020108, 10.3389/fnins.2020.00424, lee2016training}.  $T$ of \cite{wu2019direct} in Tab.~\ref{tab:previous stoa T} of the main text for N-MNIST and CIFAR10-DVS are calculated manually according to their paper. Specifically, they illustrate that the time resolution is reduced by accumulating the spike train within every 5 $ms$ and the time range ($us$) of N-MNIST and CIFAR10-DVS are [290901, 315348] and [1149758, 1459301], respectively.

\section{Hyper-Parameters}
We use the Adam \cite{kingma2014adam} optimizer with the learning rate 0.001 and the cosine annealing learning rate schedule \cite{loshchilov2016sgdr} with $T_{schedule}=64$. The \textit{batch size} is set to 16 to reduce memory consumption. The drop probability $p$ for dropout layers is 0.5. The clamp function for PLIF neurons is $k(a) = \frac{1}{1 + e^{-a}}$ and the surrogate gradient function is $\sigma(x) = \frac{1}{\pi} \arctan(\pi x) + \frac{1}{2}$, thus $\sigma'(x) = \frac{1}{1 + (\pi x)^2}$. We set $V_{reset} = 0$ and $V_{th} = 1$ for all neurons. We notice that some previous works, e.g., \cite{wu2018STBP}, \cite{wu2019direct}, fine tuned $V_{th}$ for different tasks, which is unnecessary. To be specific, as $\Theta(V - V_{th}) = \Theta(V_{th}(\frac{V}{V_{th}} - 1)) = \Theta(\frac{V}{V_{th}} - 1)$ and $V$ is directed influenced by trainable weights, setting $V_{th}=1$ implements an implicit normalization for weights, which can mitigate the exploding and vanishing gradient problem. As discovered by Zenke and Vogels \cite{Zenke2020.06.29.176925},  ignoring the neuronal reset when computing gradients by detaching them from the computational graph can improve performance, we also detach $S_{t}$ in the neuronal reset.

\section{Accuracy with a Validation Set}
The performance comparison in Tab.~\ref{tab:ACC CMP} of the main text is obtained by training on the training set, testing on the test set alternately, and recording the maximum test accuracy. Both this paper and the state-of-the-art methods use this way to report performance. However, this kind of accuracy is overestimated. Here we also report the accuracy with validation, which is obtained by splitting the origin training set into a new training set and validation set, training on the new training set, testing on the validation set alternately, and recording the test accuracy on the test set only once with the model that achieved the maximum validation accuracy. We utilize 85\% samples of each class in the origin training set as the new training set and set the rest 15\% as the validation set. The accuracy with and without the validation set of proposed methods is shown at Tab.~\ref{tab:ACC VAL}.
The experiment results in Tab.~\ref{tab:ACC CMP} of the main text and Tab.~\ref{tab:ACC VAL} show that the proposed method outperforms the state-of-the-art accuracy on nearly all datasets.
\begin{table}
	\centering
	
	\begin{tabular}{ccc}
		\toprule
		\textbf{Dataset} & \textbf{Without Validation} & \textbf{15\% Validation} \\
		\midrule
		MNIST & 99.72\% & 99.63\% \\
		Fashion-MNIST & 94.38\% & 93.85\% \\
		CIFAR-10 & 93.50\% & 92.58\% \\
		N-MNIST & 99.61\% & 99.57\% \\
		CIFAR10-DVS & 74.80\% & 69.00\% \\
		DVS128 Gesture & 97.57\% & 96.53\% \\
		\bottomrule 
	\end{tabular}
	
	\caption{Accuracy of the proposed method with/without the validation set on different datasets.}
	
	\label{tab:ACC VAL}
\end{table}

\section{Distribution of the First $\frac{W_{fc}}{\tau}$}
\begin{figure}
	\centering
	\includegraphics[width=0.5\textwidth,trim=0 0 0 0,clip]{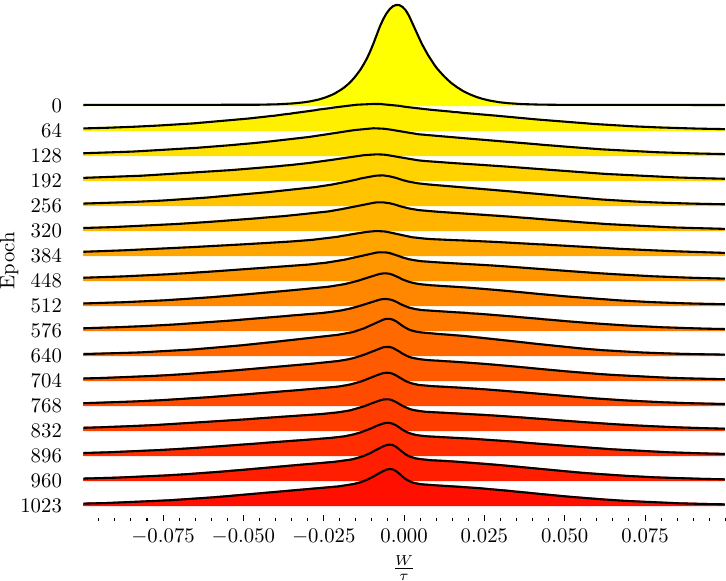}
	\caption{The distribution of $\frac{W_{fc}}{\tau}$ during training the SNN on CIFAR10-DVS.}
	\label{figure:dis}	
\end{figure}

In Sec.~\ref{Ablation Study} of the main text, we find that PLIF neurons after the first FC layer are learning to become the Non-Leaky-Integrate-and-Fire neurons as $\frac{1}{\tau} \rightarrow 0$ and $\frac{W_{fc}}{\tau}$ converges. To illustrate the convergence, we show the distribution of $\frac{W_{fc}}{\tau}$ during training the SNN on CIFAR-10DVS in Fig.~\ref{figure:dis}. The distribution of $\frac{W_{fc}}{\tau}$ on other datasets converges in the same way.

\begin{figure}
	\centering
	\includegraphics[width=0.5\textwidth,trim=50 100 40 100,clip]{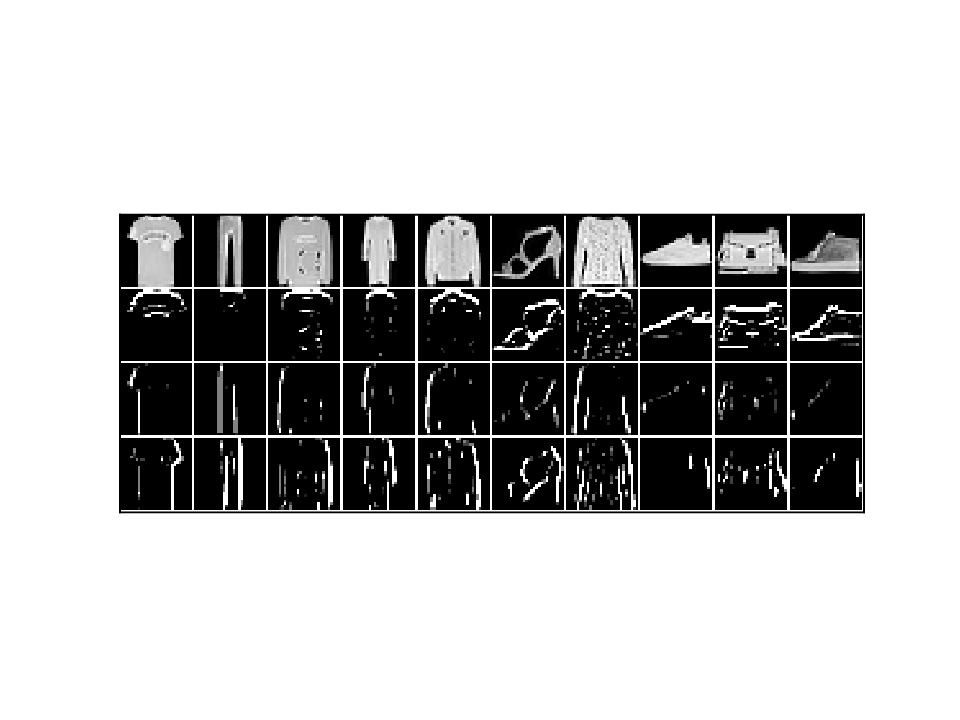}
	\caption{Ten samples from the Fashion-MNIST dataset and the corresponding firing rates $\boldsymbol F_{T_{s}=8}^{2}$ from channel 45,75 and 76 $(c=45, 75, 76)$ of the first PLIF neurons layer are shown in row 1-4. Each column represents a sample and corresponding firing rates.}
	\label{figure:spikes a1}	
\end{figure}

\begin{figure*}
	\centering
	\subfigure[$\boldsymbol S_{t=0}^{2}(c=0,1,...,127)$]{\includegraphics[width=0.45\textwidth,trim=50 70 40 86,clip]{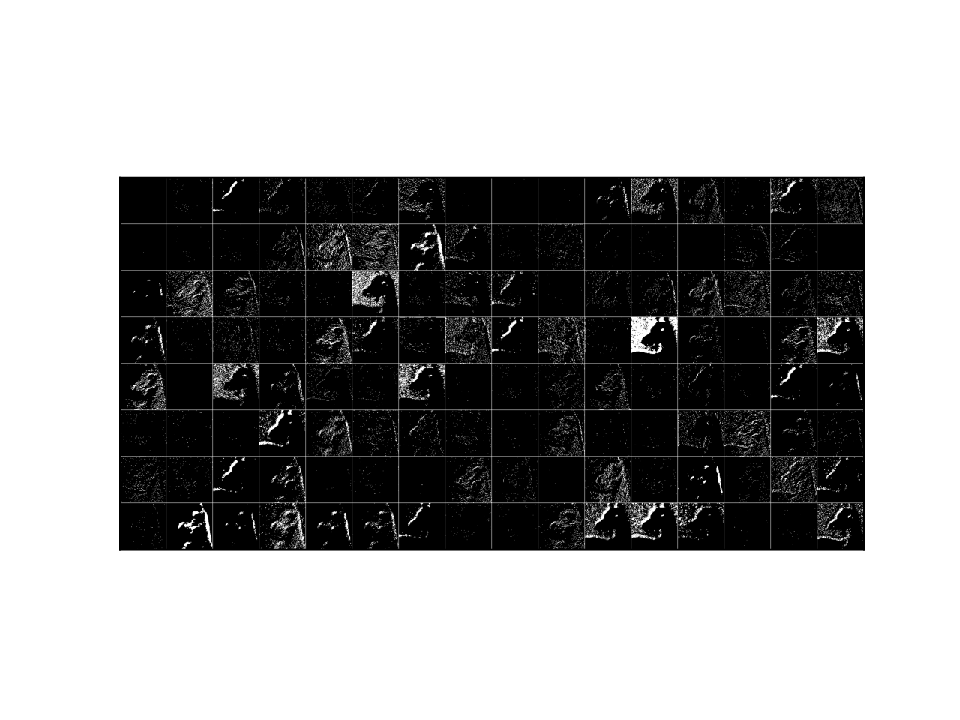}\label{figure:spikes b}}
	\subfigure[$\boldsymbol F_{T_{s}=19}^{2}(c=0,1,...,127)$]{\includegraphics[width=0.45\textwidth,trim=50 70 40 86,clip]{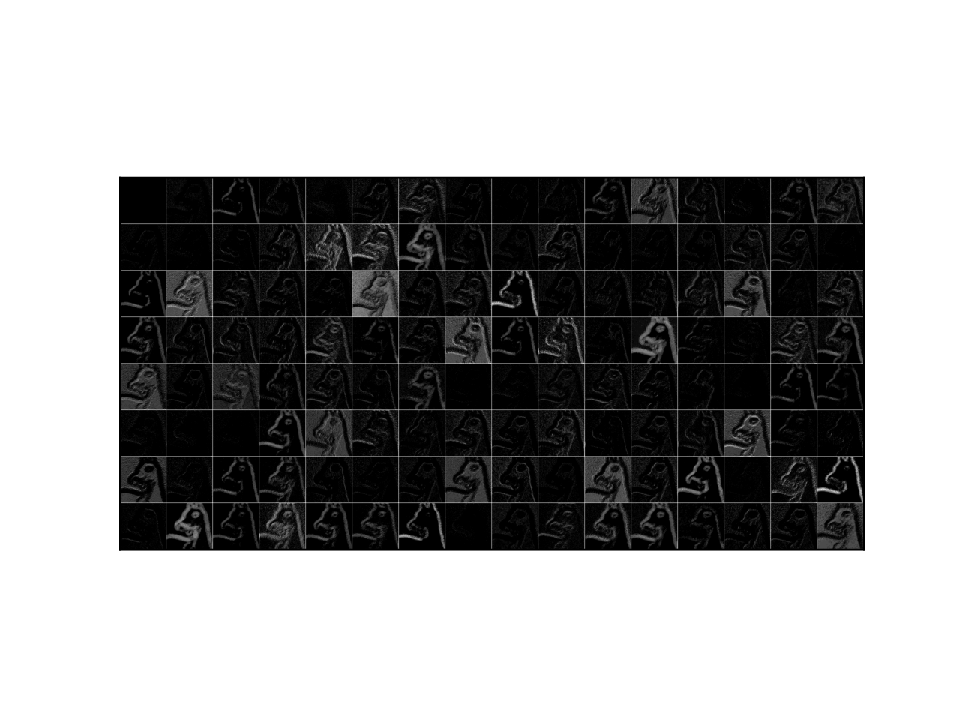}\label{figure:spikes c}}
	\caption{Given a sample labeled \textit{horse} from CIFAR10-DVS, (a) shows spikes from all 128 channels of the first spiking neurons layer at $t=0$, and (b) shows firing rates of these neurons at $T_{s}=19$.}\label{figs2}
\end{figure*}
\begin{figure*}
	\centering
	\subfigure[$\boldsymbol x_{t}$ and $\boldsymbol S_{t}^{2}(c=40, 103)$ at $t=0,1,...,19$]{\includegraphics[width=0.45\textwidth,trim=50 140 40 140,clip]{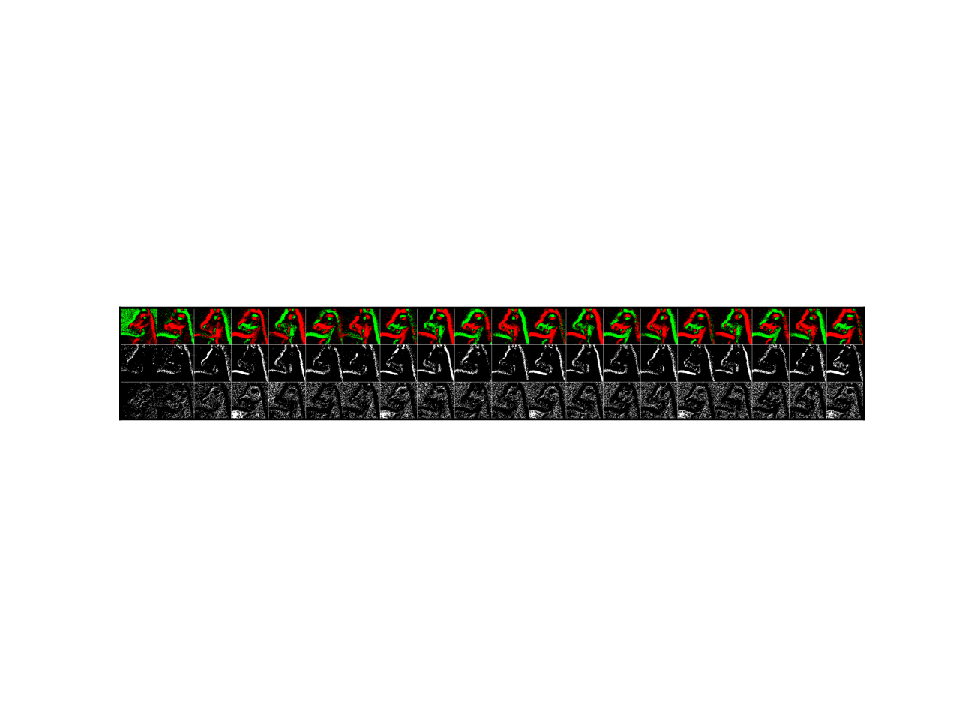}\label{figure:spikes d}}
	\subfigure[$\boldsymbol x(T_{s})$ and $\boldsymbol F_{T_{s}}^{2}(c=40, 103)$ at $T_{s}=0,1,...,19$]{\includegraphics[width=0.45\textwidth,trim=50 140 40 140,clip]{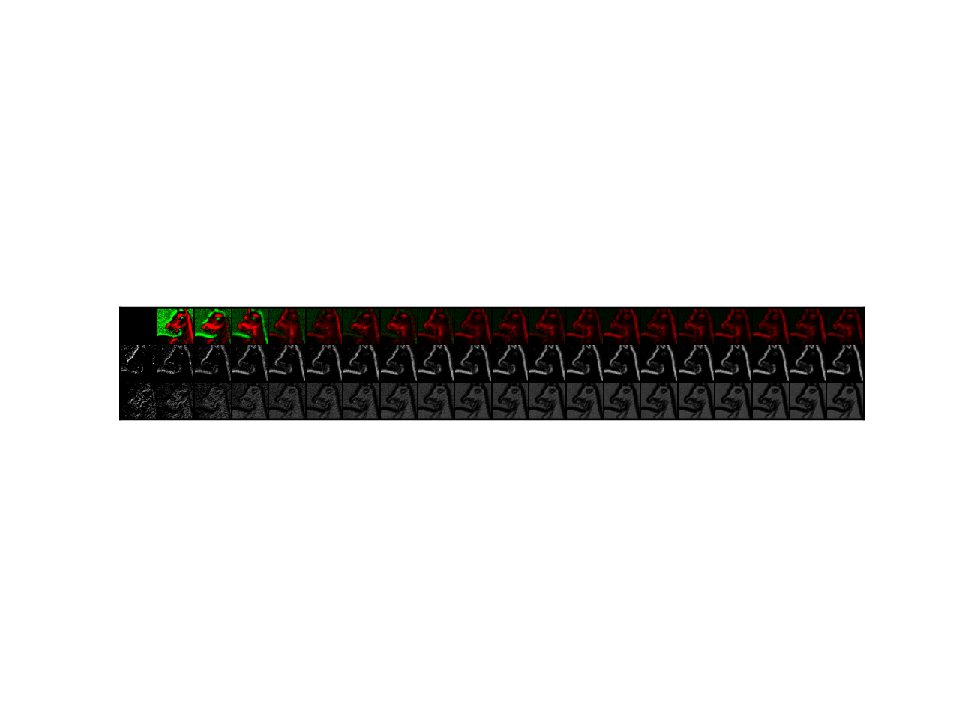}\label{figure:spikes e}}
	\caption{Given the sample sample as Fig.~\ref{figs2}, the input data and output spikes of channel 40 and 103 at each time-step are showed in (a) at row 1, 2, 3, respectively. The mean input data and firing rates of channel 40 and 103 at each time-step are showed in (b).}
\end{figure*}

\begin{figure}
	\begin{center}
		\includegraphics[width=0.5\textwidth,trim=52 110 40 120,clip]{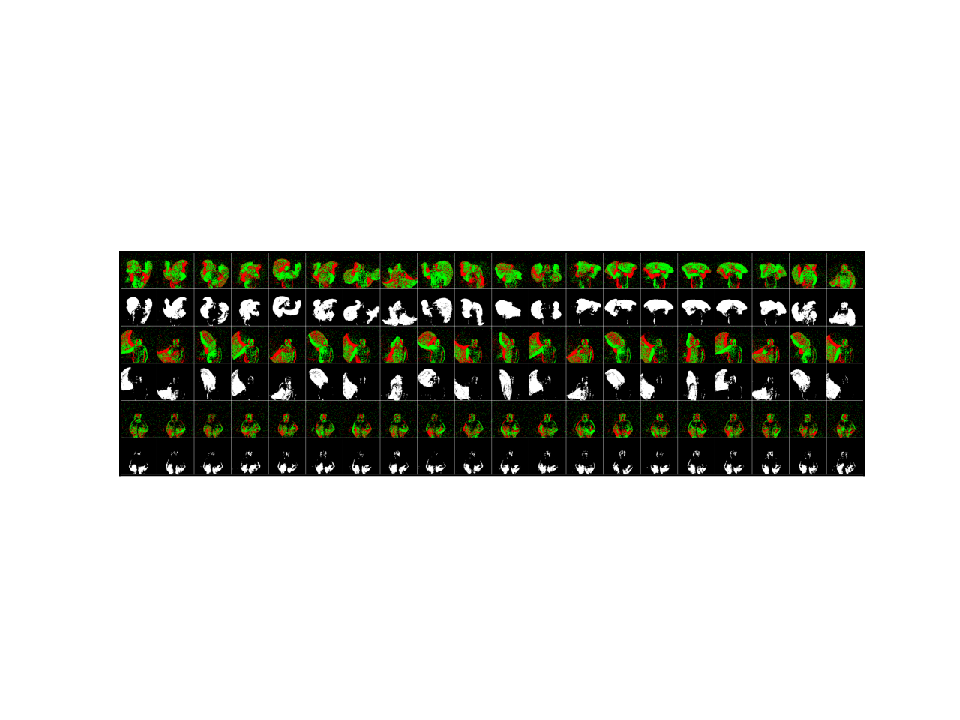}\
		\caption{Three samples from the DVS128 Gesture dataset labeled \textit{random other gestures}, \textit{right hand clockwise}, \textit{drums} are shown in row 1, 3, 5. The corresponding output spikes from channel 59 of the first PLIF neurons layer are shown in row 2, 4, 6.} 
		\label{figure:spiking encoding}
	\end{center}
	\vspace{-0.50cm}
\end{figure}

\section{Visualization of Spiking Encoder}
To evaluate the learnable encoder, we give inputs $\boldsymbol x_{t}$ to the trained network and show the output spikes $\boldsymbol S_{t}^{n}(c)$ and the firing rates $\boldsymbol F_{T_{s}}^{n}(c) = \frac{1}{T_{s}}\sum_{t=0}^{T_{s}-1}\boldsymbol S_{t}^{n}(c)$ from channel $c$ in the $n$-th layer, which is similar to \cite{DENG2020294}. Although the output spikes from deeper spiking neurons layers contain more semantic features, they are harder to read and understand. Thus we only show the spikes from the first spiking neurons layer, that is, $n = 2$.

Fig.~\ref{figure:spikes a1} illustrates 10 input images from static Fashion-MNIST dataset (row 1) and the firing rates $\boldsymbol F_{T_{s}=8}^{2}$ of three typical channel (45, 75 and 76) of the first PLIF neurons layer (row 2, 3 and 4). One can find that the firing rates from channel 45, 75 and 76 detect upper, left, right edges of the input images. Fig.~{\color{red}S}\ref{figure:spikes b} shows a 2-D grid flatten across channels from the 3-D tensor $\boldsymbol S_{t=0}^{2}(c=0,1,...,127)$ when given an input sample labeled \textit{horse}, which illustrates the features extracted by the spiking encoder at $t=0$. As the CIFAR10-DVS dataset is converted from the static CIFAR-10 dataset, the firing rates accumulated from spikes can reconstruct the images filtered by the convolutional layer. Fig.~{\color{red}S}\ref{figure:spikes c} illustrates the firing rates $\boldsymbol F_{T_{s}=19}^{2}$ of all 128 channels $(c=0,1,...,127)$, which have clearer texture than binary output spikes in Fig.~{\color{red}S}\ref{figure:spikes b}. Fig.~{\color{red}S}\ref{figure:spikes d} shows the input $\boldsymbol x_{t}$ (row 1) and the corresponding output spikes $\boldsymbol S_{t}^{2}$ of channel 40 and 103 (row 2 and 3) at $t=0, 1, ..., 19$, and Fig.~{\color{red}S}\ref{figure:spikes e} shows the mean input $\boldsymbol x(T_{s}) = \frac{1}{T_{s}}\sum_{t=0}^{T_{s}-1}\boldsymbol x_{t}$  (row 1) and the corresponding firing rates  $\boldsymbol F_{T_{s}}^{2}$ of channel 40 and 103 (row 2 and 3) at $T_{s}=0,1,...,19$. One can find that as $T_{s}$ increases, the texture constructed by firing rates $\boldsymbol F_{T_{s}}^{2}$ becomes more distinct, which is similar to the use of the Poisson encoder.

Fig.~\ref{figure:spiking encoding} visualizes three input samples $\boldsymbol x_{t}$ and output spikes  $\boldsymbol S_{t}^{2}(c=59)$ in the DVS128 Gesture dataset. Three samples labeled \textit{random other gestures}, \textit{right hand clockwise}, \textit{drums} at $t=0,1,...,19$ from the DVS128 Gesture dataset are shown in row 1, 3, 5 of Fig.~\ref{figure:spiking encoding}. For comparison, the corresponding output spikes from channel 59 of the PLIF neurons in the first conventional layer are shown in rows 2, 4, 6. One crucial difference is that the output almost only includes the gesture's response spikes, indicating that the spiking neurons implement efficient and accurate filtering on both spatial-variant and temporal-variant input data, reserving the gesture but discarding the player.

\section{Relations between different Encoders}
The Poisson encoder is one of the rate encoding methods and widely used in SNNs \cite{diehl2015unsupervised,lee2020enabling,shrestha2018slayer, ST-RSBP, LISNN, Han_2020_CVPR} to encode images into spikes. Given a image pixel $p \in[0, 1]$, the encoded spike $S_{t}$ at time-step $t$ is fired with the probability $p$. Thus, the expectation of the number of spikes during the whole time-steps $T$ is $E_{Poisson}(\Sigma_{t=0}^{T-1} S_{t}) = pT$. In our poposed SNNs, the input is directly fed to the network without being first converted to spikes and the image-spike encoding is done by the first \textit{\{Conv2d-Spiking Neurons\}} module ($BN$ is omitted), which can be seen as a learnable encoder. Here we denote this encoder as $ENC_{l}$. If we set $Conv2d$ non-learnable with $channels=kernel~size=1$, the kernel weight as the constant $w > 0$, and $Spiking~Neurons$ as Non-Leaky-Integrate-and-Fire neurons with threshold potential $V_{th}$ and $V_{reset}=0$, then the expectation of spikes number of this module is $E(\Sigma_{t=0}^{T-1} S_{t}) = \lfloor \frac{T}{\lceil \frac{V_{th}}{wp} \rceil} \rfloor$, where $\lceil \rceil$ denotes the ceiling operation. We can find that $E(\Sigma_{t=0}^{T-1} S_{t}) \approx E_{Poisson}(\Sigma_{t=0}^{T-1} S_{t})$ when $V_{th} = w = 1$, which indicates that $ENC_{l}$ can approximate the function of the Possion encoder in rate encoding.

The latency encoder used in \cite{WYSOSKI20082563, Yu2012PatternRC, 6469239, YU20143} is a representative temporal encoding method. The latency encoder encodes the image pixel $p$ into a spike at time-step $t_{p}$. Thus, the information of input is encoded in the precise firing time of the spike. $t_{p}$ is usually inversely proportional to the input intensity $p$, e.g., $t_{p} = \lfloor (T_{max} - 1) (1 - p) \rfloor$ and $T_{max}$ is the encoding period. We can also find that the first firing time of $ENC_{l}$ for the given input $p$ is $\lceil \frac{V_{th}}{wp} \rceil$, which satisfies that the larger input intensity $p$ causes the faster spike. In fact, the latency encoder is an extremely simplified learnable encoder with directly inputted images. In this paper, the proposed learnable encoders have learnable weights and more channels, which is able to encode images into complex spikes pattern with more semantic information, e.g., reserving the gesture but discarding the player of samples from DVS128 Gesture dataset (see Fig.~\ref{figure:spiking encoding}).


%
%
%
%
%
%
\end{appendices}

{\small
	\bibliographystyle{ieee_fullname}
	\bibliography{egbib}
}

\end{document}


\title{Supplementary Materials for: Incorporating Learnable Membrane Time Constant to Enhance Learning of Spiking Neural Networks}

\author{Wei Fang$^{1,2}$ 
	\and Zhaofei Yu$^{1,2}$ 
	\and Yanqi Chen$^{1,2}$ 
	\and Timothée Masquelier $^{3}$ 
	\and Tiejun Huang$^{1,2}$
	\and Yonghong Tian$^{1,2}$\\
	$^1$Department of Computer Science and Technology, Peking University\\
	$^2$Peng Cheng Laboratory, Shenzhen 518055, China\\
	$^3$Centre de Recherche Cerveau et Cognition (CERCO), UMR5549 CNRS - Univ. Toulouse 3 , Toulouse, France\\
	{\tt\small 
		\{fwei, yuzf12, chyq\}@pku.edu.cn,
		timothee.masquelier@cnrs.fr,
		\{tjhuang, yhtian\}@pku.edu.cn
	}
}
\maketitle
\section{Reproducibility}
All the experiments are implemented by *** \footnote{\url{***}}, which is an open-source deep learning framework for SNNs based on PyTorch \cite{PYTORCH}. All of the source codes, training logs, and trained models are available at \url{***}. To maximize reproducibility, we use identical seed in all codes.

\section{RNN-like Expression of LIF Neuron}
Eq.~(\ref{LIF iterative expression}) can be written as
\begin{align}
	V_{t} = \left(1 - \frac{1}{\tau}\right)V_{t-1} + \frac{1}{\tau}WX_{t}
	\label{LIF iterative expression 2}
\end{align}

The integration progress $\frac{1}{\tau}WX_{t}$ makes the LIF neuron be able to remember current input information, while the leakage progress $(1 - \frac{1}{\tau})V_{t-1}$ can be seen as forgetting past information. Eq.~(\ref{LIF iterative expression 2}) shows that the balance between remembering and forgetting is controlled by $\tau$, which plays an analogous role with the gates in LSTM (Long Short-Term Memory) \cite{hochreiter1997long}.

\section{Datasets Introduction}
\paragraph{MNIST}
The MNIST database of handwritten digits is composed of $28 \times 28$ gray-scale images and labeled from 0 to 9. The MNIST dataset has 60,000 training images and 10,000 test images.
\paragraph{Fashion-MNIST}
The Fashion-MNIST dataset is similar to the MNIST dataset consisting of a training set of 60,000 examples and a test set of 10,000 examples. Each example in the Fashion-MNIST dataset is also a $28 \times 28$ gray-scale image with a label from 10 classes.
\paragraph{CIFAR-10}
The CIFAR-10 dataset consists 60,000 natural images in 10 classes, with 6000 images per class. The number of training images is 50,000 and that of test images is 10,000.
\paragraph{N-MNIST}
The Neuromorphic-MNIST (N-MNIST) dataset is a spiking version of the MNIST dataset by the neuromorphic sensor. It was converted from MNIST by mounting the ATIS sensor on a motorized pan-tilt unit and having the sensor move while it views MNIST examples on an LCD monitor as shown in this video. It consists of 60,000 training examples and 10,000 test examples.
\paragraph{CIFAR10-DVS}
The CIFAR10-DVS dataset is the neuromorphic vision dataset of the CIFAR-10 dataset. It is composed of 10,000 examples in 10 classes, with 1000 examples in each class.
\paragraph{DVS128 Gesture}
The DVS128 Gesture dataset contains 11 hand gestures from 29 subjects under 3 illumination conditions recorded by a DVS128 camera.
\begin{table}
	\centering
	\begin{tabular}{lp{5.2cm}p{0.8cm}}
		\hline
		Dataset & Network Structure\\
		\hline
		\makecell[l]{MNIST \\ Fashion-MNIST \\ N-MNIST} & \makecell[l]{\{c128k3s1-BN-PLIF-MPk2s2\}*2-\\DP-FC2048-PLIF-DP-FC100-PLIF-\\APk10s10}\\
		
		CIFAR-10 & \{\{c256k3s1-BN-PLIF\}*3-MPk2s2\}*2-DP-FC2048-PLIF-DP-FC100-PLIF-APk10s10 \\
		
		CIFAR10-DVS &
		\{c128k3s1-BN-PLIF-MPk2s2\}*4-DP-FC512-DP-PLIF-FC100-PLIF-APk10s10 \\
		
		DVS128 Gesture &
		\{c128k3s1-BN-PLIF-MPk2s2\}*5-DP-FC512-DP-PLIF-FC110-PLIF-APk10s10 \\
		\hline
	\end{tabular}
	\caption{Network structures for different datasets.}
	\label{tab:detailed network structures}
\end{table}

\section{Network Structure Details}
Tab.\,\ref{tab:detailed network structures} gives network structure details. \textit{c128k3s1} represents the convolutional layer with output channels 128, kernel size 3 and stride 1. \textit{BN} is the batch normalization \cite{BN}. \textit{MPk2s2} is the spike max pooling layer with kernel size 2 and stride 2. \textit{PLIF} is the parametric LIF spiking neuron layer. \textit{DP} represents the time-invariant dropout proposed in \cite{lee2020enabling}. \textit{FC2014} represents the fully connected layer with output features 2048. The symbol \textit{\{\}*} indicates the repeated structure. For example, \textit{\{c128k3s1-BN-PLIF-MPk2s2\}*2} means that there are two \textit{c128k3s1-BN-PLIF-MPk2s2} modules connected sequentially. We use a vote layer to boost classifying robustness. The vote layer is implemented by an average pool with kernel size $M=10$ and stride $M=10$, e.g. \textit{APk10s10}, which uses $M$ neurons' output spikes to represent one class and these $M$ neurons can be seen as a neuron population.

\section{Training Algorithm to Fit Target Output}
After defining the derivative of spiking function $\Theta(x)$, the parameters of SNNs can be trained by gradient descent as that in ANNs. Image classification, which is the task in this article, and other tasks for both ANNs and SNNs can be seen as optimizing parameters of network to fit a target output when given a certain input. The training algorithm for SNNs we used can be described as followed:
\begin{algorithm}[H]
	\caption{Gradient Descent Algorithm for SNNs to Fit Target Output}
	\textbf{Require:} learning rate $\epsilon$, network's parameter $\boldsymbol \theta$, simulating step $T$, input $\boldsymbol X = \{\boldsymbol X_{0}, \boldsymbol X_{1}, ..., \boldsymbol X_{T-1}\}$, target output $\boldsymbol Y = \{\boldsymbol Y_{0}, \boldsymbol Y_{1}, ..., \boldsymbol Y_{T-1}\}$, loss function $L=\mathcal L(\boldsymbol Y, \boldsymbol O)$
	
	initialize $\boldsymbol \theta$
	
	create an empty list $\boldsymbol S = \{ \}$
	
	\algorithmicfor{ $t \gets 0, 1, ... T-1$}
	
	~~input $\boldsymbol X_{t}$ to network, get output spike $\boldsymbol S_{t}$
	
	~~append $\boldsymbol S_{t}$ to $\boldsymbol S = \{\boldsymbol S_{0}, \boldsymbol S_{1}, ..., \boldsymbol S_{t-1}\}$
	
	calculate loss $L=\mathcal L(\boldsymbol Y, \boldsymbol O)$
	
	update parameter $\boldsymbol \theta =\boldsymbol \theta - \epsilon \cdot \bigtriangledown_{\boldsymbol \theta} L$
\end{algorithm}

The loss function $L=\mathcal L(\boldsymbol Y, \boldsymbol O)$ should be a distance measure between $\boldsymbol Y$ and $\boldsymbol S$.

\section{Hyper-parameter}
We use Adam \cite{kingma2014adam} optimizer with learning rate 0.001 and cosine annealing learning rate schedule \cite{loshchilov2016sgdr} with $T_{schedule} = 64$. The batch size is set to 16 to reduce memory consumption. We set $\tau_{0} = 2$ for all PLIF neurons. The drop probability $p$ for dropout layers is 0.5, the clamp function for $\tau$ is $k(a) = \frac{1}{1 + e^{-x}}$ and the surrogate gradient function is $\sigma(x) = \frac{1}{\pi} \arctan(\pi x) + \frac{1}{2}$. We set $V_{reset}=0$ and $V_{th}=1$ for all neurons. We notice that some previous works, e.g., \cite{wu2018STBP}, \cite{wu2019direct}, finetuned $V_{th}$ for different tasks, which we regard unnecessary because $\Theta(V - V_{th}) = \Theta(V_{th}(\frac{V}{V_{th}} - 1)) = \Theta(\frac{V}{V_{th}} - 1)$ and $V$ is directed influenced by trainable weights. Setting $V_{th}=1$ implements an implicit normalization for weights, which may decrease the gradient vanish or explode. It is worth noting that \cite{Zenke2020.06.29.176925} discovers that ignoring the spike reset when computing gradients by detaching it from the computational graph can improve performance. Thus we also detach $S_{t}$ in Eq.~(\ref{neural reset}). Note that CIFAR10-DVS dataset doesn't provide division for training and testing. We choose the first 9000 samples for training and the rest 1000 samples for testing in each class, which is similar to \cite{wu2019direct}.

\section{Visualization of Spiking Encoding}
As we mentioned above, the convolutional layers with spiking neurons can be seen as a spiking encoder that extracts features from the analog input and converts them to spikes. Different from the Poisson encoder, which is widely used in SNNs such as \cite{diehl2015unsupervised}, \cite{lee2020enabling}, \cite{shrestha2018slayer}, this spiking encoder is a learnable encoder. To evaluate the encoder, we give inputs $x_{t, c}$ to the trained network and show output spikes $y_{n}(t, c)$ and firing rates $Y_{n}(T, c) = \sum_{t=0}^{T-1}y_{n}(t, c) / T$ in the $n$-th layer in temporal domain, which is similar to \cite{DENG2020294}. Output spikes from deeper spiking neurons layers contain more semantic features but are harder for humans to read and understand, so we only show the visualization from the first spiking neurons layer, where $n = 2$.

Fig.\,\ref{figure:spikes a} shows 10 input images from static Fashion-MNIST dataset at the first row and $Y_{2}(T=8, c=45), Y_{2}(T=8, c=75), Y_{2}(T=8, c=76)$ at the three sequential rows. $c=15, 75, 76$ represent three typical functions of the spiking convolutional layers that detect upper, left, right edges from the input images. Fig.\,\ref{figure:spikes b} shows a 2-D grid flatten across channels from the 3-D tensor $y_{2}(t=0)$ when given an input sample labeled \textit{horse}, which illustrates the features extracted by the spiking encoder at $t=0$. Note that CIFAR10-DVS is converted from the static CIFAR-10, and the firing rates which are accumulated from spikes can reconstructed the images filtered by the convolutional layer. Fig.\,\ref{figure:spikes c} is the firing rates $Y_{2}(T=19)$, showing clearer texture than binary Fig.\,\ref{figure:spikes b}. Fig.\,\ref{figure:spikes d} shows $x(t)$ and $y_{2}(t, c=40, 103)$ at $t=0, 1, ..., 19$ and Fig.\,\ref{figure:spikes e} shows the accumulated [$X(T) = \sum_{t=0}^{T-1}x(t)$ and $Y_{2}(T, c=40, 103)$. Fig.\,\ref{figure:spikes e} indicates $Y_{2}(T, c=40, 103)$ becomes more distinct with the increment of $T$, which is similar with using the possion encoder. 

\begin{figure}[H]
	\centering
	\subfigure[Ten samples from Fashion-MNIST and corresponding $Y_{2}(T=8, c=45, 75, 76)$]{\includegraphics[width=0.5\textwidth,trim=50 100 40 100,clip]{./fig/spikes/fmnist/input_firing_rate__45_75_76.eps}\label{figure:spikes a}}
	
	\subfigure[$y_{2}(t=0)$ for a given sample labeled \textit{horse} from CIFAR10-DVS]{\includegraphics[width=0.5\textwidth,trim=50 70 40 86,clip]{./fig/spikes/cf10dvs/3_0.eps}\label{figure:spikes b}}
	
	\subfigure[$Y_{2}(T=19)$ for a given sample labeled \textit{horse} from CIFAR10-DVS]{\includegraphics[width=0.5\textwidth,trim=50 70 40 86,clip]{./fig/spikes/cf10dvs/3_sum_19.eps}\label{figure:spikes c}}
\end{figure}

\begin{figure}[H]
	\subfigure[$x(t)$ from CIFAR10-DVS and $y_{2}(t, c=40, 103)$ at $t=0,1,...,19$]{\includegraphics[width=0.5\textwidth,trim=50 140 40 140,clip]{./fig/spikes/cf10dvs/map_idx__40_103.eps}\label{figure:spikes d}}
	
	\subfigure[$X(T) = \sum_{t=0}^{T-1}x(t)$ and $Y_{2}(T, c=40, 103)$ at $t=0,1,...,19$]{\includegraphics[width=0.5\textwidth,trim=50 140 40 140,clip]{./fig/spikes/cf10dvs/map_sum_idx__40_103.eps}\label{figure:spikes e}}
	
	\caption{Visualization of spike encoding in our networks. 
		Each column in (a) represents a sample from Fashion MNIST and firing rates from the first spiking neurons layer, showing $c=15, 75, 76$ detect upper, left, right edges from the input images. 
		(b) shows spikes from all channels of the first spiking neurons layer at $t=0$ when given a sample labeled \textit{horse} from CIFAR10-DVS and (c) shows firing rates of these neurons at $T=19$. (d) and (e) show the input data, output spikes, and the accumulated input and output, respectively. (b-e) indicate that the spiking encoding in the first 3 layer extract features from the input data and the extracted information become more distinct with the increment of $T$.}
\end{figure}